\documentclass[sigconf]{acmart}

\AtBeginDocument{%
  }

\usepackage{booktabs}
\usepackage{amsmath}
\usepackage{graphicx}
\usepackage{multirow}
\usepackage{hyperref}
\usepackage{balance}
\usepackage{enumitem}
\usepackage[table]{xcolor}
\usepackage{array}
\usepackage{colortbl}
\usepackage{placeins}

\definecolor{lightgrayA}{RGB}{242,242,242}
\definecolor{lightgrayB}{RGB}{248,248,248}
\definecolor{rulegray}{RGB}{180,180,180}

\definecolor{headergray}{RGB}{235,237,240}
\definecolor{groupgray}{RGB}{246,247,249}
\definecolor{adeblue}{RGB}{232,243,252}
\definecolor{fdeorange}{RGB}{252,241,229}

\definecolor{tablesubheader}{RGB}{244,246,248}
\definecolor{tabletotal}{RGB}{230,241,250}

\definecolor{tableheader}{RGB}{232,236,240}
\definecolor{tablegroup}{RGB}{246,247,249}
\definecolor{tableours}{RGB}{232,243,252}

\definecolor{tabgray}{RGB}{245,245,245}
\definecolor{avgblue}{RGB}{232,244,252}
\newcolumntype{G}{>{\columncolor{tabgray}}c}
\newcolumntype{B}{>{\columncolor{avgblue}}c}

\setcopyright{none}
\acmConference[KDD]{Proceedings of the ACM SIGKDD Conference on Knowledge Discovery and Data Mining (Datasets \& Benchmarks Track)}{2027}{}
\acmBooktitle{Proceedings of the ACM SIGKDD Conference on Knowledge Discovery and Data Mining}
\acmPrice{}
\acmDOI{}
\acmISBN{}
\renewcommand\footnotetextcopyrightpermission[1]{}

\makeatletter
\if@ACM@anonymous
  \newcommand{\repolink}{an anonymized repository provided in the supplementary material}
  \newcommand{\datalink}{an anonymized archive provided in the supplementary material}

\else
  \newcommand{\repolink}{\url{https://github.com/mark000071/Envship_Framework_vessel_traj_pred}}
  \newcommand{\datalink}{\url{https://huggingface.co/datasets/mark000071/envship_v2_datasets}}

\fi
\makeatother

\title{EnvShip: A Unified Framework for Context-Aware and Cross-Region Vessel Trajectory Forecasting}

\renewcommand{\shortauthors}{Ma et al.}

\author{Kun Ma}
\affiliation{%
  \institution{Harbin Engineering University}
  \city{Harbin}
  \country{China}}
\affiliation{%
  \institution{Politecnico di Torino}
  \city{Turin}
  \country{Italy}}

\author{Qilong Han}
\affiliation{%
  \institution{Harbin Engineering University}
  \city{Harbin}
  \country{China}}

\author{Jingzheng Yao}
\affiliation{%
  \institution{Harbin Engineering University}
  \city{Harbin}
  \country{China}}

\author{Chengjing Song}
\affiliation{%
  \institution{Harbin Engineering University}
  \city{Harbin}
  \country{China}}

\author{Hao Wang}
\affiliation{%
  \institution{Harbin Engineering University}
  \city{Harbin}
  \country{China}}

\author{Changmao Wu}
\affiliation{%
  \institution{
   Chinese Academy of Sciences}
  \city{Beijing}
  \country{China}}

\author{Carla Fabiana Chiasserini}
\affiliation{%
  \institution{Politecnico di Torino}
  \city{Turin}
  \country{Italy}}
\begin{document}

\begin{abstract}
Accurate vessel trajectory forecasting is essential for maritime situational awareness, navigation safety, traffic management, and autonomous navigation. Public Automatic Identification System (AIS) archives have enabled extensive research in this area, yet results remain difficult to compare because existing studies use incompatible preprocessing pipelines, forecasting horizons, data splits, coordinate systems, contextual inputs, and evaluation settings. We present EnvShip, a unified multi-region framework for context-aware and cross-region vessel trajectory forecasting. EnvShip applies a fixed and reproducible pipeline to public AIS data from Denmark, the United States, Greece, and Norway, and defines two standardized forecasting tracks spanning short- and long-horizon settings. From large-scale processed data, we curate 330{,}000 short-term and 106{,}857 long-horizon samples through strict motion screening, vessel-category and difficulty stratification, redundancy control, and vessel-disjoint splits. Each sample is aligned with environmental and neighboring-vessel context, together with weather and sea-state variables where available. We evaluate representative methods under in-domain and cross-region protocols, with analyses across prediction difficulty, scene type, and random seeds. Results show that environmental context provides the largest gains in coastline-constrained scenes, whereas neighboring-vessel context primarily benefits interaction-rich cases. Multi-region training improves generalization in most settings but introduces negative transfer for some source combinations. EnvShip provides a common and reproducible testbed for vessel trajectory forecasting.

\end{abstract}

\ccsdesc[500]{Computing methodologies~Tracking}
\ccsdesc[300]{Information systems~Data mining}

\keywords{Vessel trajectory prediction; AIS dataset; context-aware maritime benchmark; cross-domain transfer
}

\maketitle
\pagestyle{plain}

\section{Introduction}
\begin{figure*}[t]
    \centering
    \includegraphics[width=1\textwidth]{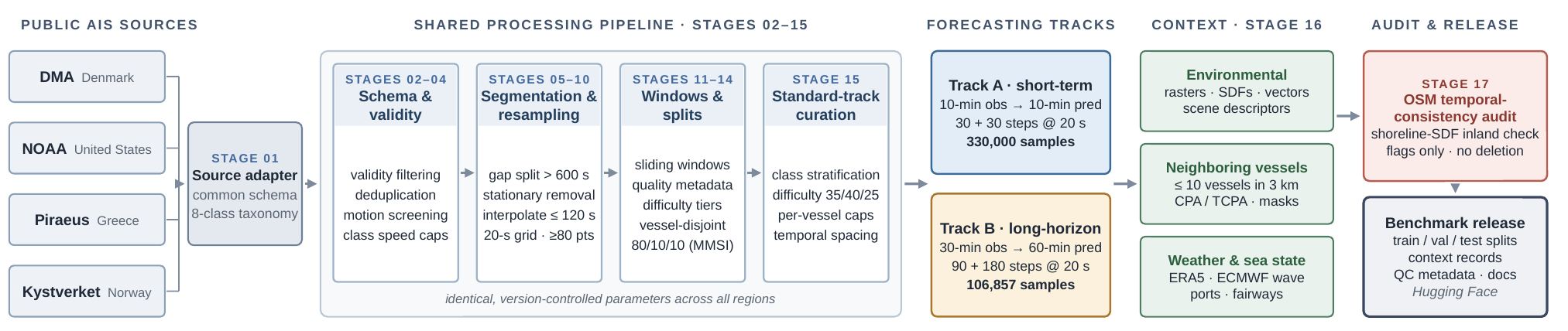}
    \caption{\textbf{EnvShip construction pipeline.}
Four public AIS sources are processed through a shared pipeline to produce two forecasting tracks with aligned context and quality-control metadata.}
    \label{fig:pipeline}
\end{figure*}
Maritime transport carries more than 80\% of international trade in goods by volume, making safe and efficient vessel operations important to global supply chains~\cite{unctad2025review}. A key capability is to predict how vessels will move in the near future. Vessel trajectory forecasting supports route monitoring, collision-risk assessment, traffic management, anomaly detection, port operations, and autonomous navigation.

The Automatic Identification System (AIS) continuously reports vessel positions and motion states, and public AIS archives have enabled extensive research on data-driven vessel forecasting~\cite{AIS-zongshu-2017-8025635,li2023ship,zhang2022maritime_review,nguyen2024traisformer,qiang2023mstformer,capobianco2021deep_vessel,liu2022dl_powered}. However, results from different studies are often not directly comparable. Existing works commonly use different rules for data cleaning, trajectory segmentation, interpolation, and resampling. They also differ in observation and prediction horizons, coordinate systems, data splits, contextual inputs, and evaluation settings~\cite{zhang2018ais_reconstruction,chen2020ais_reconstruction,zhang2022ais_denoising,zhang2022maritime_review}. Consequently, an apparent improvement may reflect differences in data or task construction rather than the forecasting model itself.

Most public AIS resources are trajectory-centric and do not provide environmental or neighboring-vessel information aligned with each forecasting sample, although coastlines, navigable waterways, port geometry, and surrounding traffic can constrain vessel motion~\cite{ray2019heterogeneous,rela-GP-rong2019ship,is-gcnn-liang2022fine,gao2024multi}. Moreover, most resources cover a single country, port, or waterway~\cite{tritsarolis2022piraeus,ushant_ais_dataset,ray2019heterogeneous,dtu_abnormal_ais_dataset}. This makes it difficult to determine whether a conclusion obtained in one navigation environment remains valid in another.
To address these limitations, we present \textbf{EnvShip}, a unified multi-region framework for context-aware and cross-region vessel trajectory forecasting. EnvShip processes four public AIS sources through a fixed and reproducible pipeline: the Danish Maritime Authority archive, the NOAA MarineCadastre archive for United States waters, the Piraeus dataset from Greece, and the Kystverket feed for Norwegian coastal waters. Together, these sources cover open-water transit, coastal navigation, port operations, ferry traffic, and constrained waterways. 


EnvShip defines two standardized tracks at 20-s intervals:
Track~A uses 10 min of observation and up to 10 min of prediction,
whereas Track~B uses 30 min of observation and up to 60 min of
prediction. From large-scale regional cores, we curate 330{,}000
short-term and 106{,}857 long-horizon samples using motion
screening, vessel-category and difficulty stratification,
redundancy control, and vessel-disjoint splits.
Each forecasting sample is aligned with environmental geometry,
neighboring-vessel interactions, and weather and sea-state variables
where available. Figure~\ref{fig:pipeline} summarizes the complete construction
and release pipeline. The release also includes sample-level quality metadata and an
OpenStreetMap temporal-consistency flag for map changes that can
make historical trajectories appear inland
~\cite{haklay2008openstreetmap,openstreetmap}. 

We evaluate representative methods from four families: physics-based methods, classical machine-learning methods, trajectory-only deep models, and context-aware deep models. The evaluation covers both forecasting tracks and all four regions. Beyond aggregate in-domain results, we report performance by prediction difficulty, scene type, and neighboring-vessel density. We also conduct a multi-seed stability analysis, long-horizon evaluation, and combined and leave-one-source-out (LOSO) cross-region experiments.

The evaluation addresses three questions: (1) when contextual information improves vessel forecasting, (2) whether conclusions obtained in one region generalize to other regions, and (3) how stable model performance remains across horizons and random seeds. The results show that environmental context provides the largest gains when coastline geometry constrains vessel motion, whereas social context mainly benefits interaction-rich samples. Multi-region training generally improves robustness and helps data-scarce regions, but combining substantially different navigation regimes can introduce negative transfer. Some context-aware models are also more sensitive to random initialization than trajectory-only baselines, highlighting the importance of multi-seed evaluation.

The main contributions of this work are thus as follows:
\vspace{-0.2em}
\begin{itemize}[leftmargin=*, topsep=1pt, itemsep=1pt, parsep=0pt, partopsep=0pt]
    \item \textbf{A unified multi-region and multi-horizon framework.}
EnvShip processes public AIS data from four maritime regions through
one reproducible pipeline and defines two multi-horizon forecasting
tracks with vessel-disjoint splits, thereby preventing cross-split
leakage from overlapping sliding windows of the same vessel.

    \item \textbf{Sample-aligned context with quality control.}
Each sample is aligned with environmental, neighboring-vessel,
weather, and sea-state information, along with sample-level
quality metadata and an OpenStreetMap temporal-consistency flag.

    \item \textbf{A comprehensive forecasting evaluation.}
We evaluate representative methods under difficulty- and
scene-stratified, multi-seed, in-domain, multi-horizon, and
cross-region settings. Experimental results reveal when contextual information and
multi-region training benefit or degrade forecasting performance.
\end{itemize}
\vspace{-0.2em}

\begin{table*}[t]
\centering
\caption{Comparison of representative public AIS resources and
study-associated datasets for vessel trajectory forecasting
}
\label{tab:dataset_comparison}

\footnotesize
\setlength{\tabcolsep}{3.0pt}
\renewcommand{\arraystretch}{1.10}

\resizebox{\textwidth}{!}{%
\begin{tabular}{
p{2.85cm}
p{2.20cm}
p{2.80cm}
p{3.10cm}
p{3.90cm}
p{2.45cm}
p{5.35cm}}
\toprule

\rowcolor{tableheader}
\textbf{Resource}
& \textbf{Coverage}
& \textbf{Release / primary use}
& \textbf{Reported scale}
& \textbf{Forecast protocol}
& \textbf{Evaluation split}
& \textbf{Context support} \\
\midrule

\rowcolor{tablegroup}
\multicolumn{7}{l}{\textbf{Raw national AIS archives}} \\

DMA Historical AIS~\cite{dma_historical_ais}
& Danish waters
& Raw national AIS archive
& Multi-year monthly CSV/ZIP files
& User-defined
& User-defined
& AIS positions and motion states, vessel identifiers, type,
dimensions, and navigational status; no aligned external context \\

NOAA AccessAIS / MarineCadastre~
\cite{marinecadastre_accessais,marinecadastre_aisfaq}
& U.S. coastal and inland waters
& AIS archive and data-query service
& National-scale, multi-year terrestrial AIS records
& User-defined
& User-defined
& AIS positions and motion states, vessel identifiers, type,
dimensions, and navigational status; no aligned external context \\

\addlinespace[0.7mm]
\rowcolor{tablegroup}
\multicolumn{7}{l}{\textbf{Processed regional AIS resources}} \\

Piraeus AIS dataset~\cite{tritsarolis2022piraeus}
& Piraeus and Saronic Gulf, Greece
& Regional maritime mobility dataset
& 244M+ records over 2.5+ years
& User-defined
& User-defined
& Vessel, geographic-area, and meteorological information
provided as dataset-level tables \\

Ushant AIS dataset~\cite{ushant_ais_dataset}
& Ushant TSS, France
& Processed trajectory corpus
& 18{,}603 trajectories and 7M+ observations over 6 months
& User-defined
& User-defined
& Trajectory sequences and vessel-category metadata;
no aligned environmental or interaction context \\

Brest integrated dataset~\cite{ray2019heterogeneous}
& French Atlantic, English Channel, and Bay of Biscay
& Integrated maritime intelligence dataset
& 6 months of multi-source maritime data
& User-defined
& User-defined
& Vessel and navigation data with ports, coastlines, maritime
zones, ocean conditions, and weather at dataset level \\

\addlinespace[0.7mm]
\rowcolor{tablegroup}
\multicolumn{7}{l}{\textbf{Forecasting-oriented processed datasets}} \\

TrAISformer processed set~\cite{nguyen2024traisformer}
& Danish Straits
& Study-associated long-horizon forecasting data
& Jan.--Mar. 2019; approximately 712M raw AIS messages
& 3-h observation to up to 15-h prediction;
10-min sampling
& Study-defined temporal split
& Latitude, longitude, SOG, COG, timestamp, and vessel
identifier sequences; no external context \\

MSTFormer datasets~\cite{qiang2023mstformer}
& Gulf of Mexico
& Study-associated forecasting datasets
& 43{,}000--172{,}100 windows across four settings
& Four study-defined observation--prediction settings;
1- or 3-min sampling
& Study-defined 70/10/20 split
& AIS motion attributes and derived speed- and
course-change features; no external context \\

\midrule

\rowcolor{tableours}
\textbf{EnvShip (ours)}
& \textbf{Denmark, U.S., Greece, and Norway}
& \textbf{Unified forecasting benchmark and framework}
& \textbf{9.5M regional-core windows, 330{,}000 short-term and
106{,}857 long-horizon benchmark samples}
& \textbf{10-min observation, up to 10-min prediction;
30-min observation, up to 60-min prediction;
20-s sampling}
& \textbf{Fixed vessel-disjoint train/validation/test splits}
& \textbf{Dynamic and static AIS attributes; environmental rasters, SDFs,
shorelines, and scene descriptors; neighboring-vessel histories with
CPA/TCPA; weather and sea-state variables; and quality metadata
aligned with each sample} \\

\bottomrule
\end{tabular}%
}
\end{table*}

\section{Related Work}

\noindent\textbf{Public AIS resources and forecasting protocols.}
Public AIS data are available as raw national archives, processed regional corpora, and task-specific datasets. The Danish Maritime Authority archive~\cite{dma_historical_ais} and NOAA AccessAIS~\cite{marinecadastre_accessais,marinecadastre_aisfaq} provide large-scale point-level AIS records from which users must construct trajectories and forecasting tasks. Processed regional resources include the Piraeus dataset~\cite{tritsarolis2022piraeus}, the Ushant corpus~\cite{ushant_ais_dataset}, and the Brest datasets~\cite{ray2019heterogeneous,brest_synopses_2018}. The DTU dataset, in contrast, was developed primarily for abnormal-behavior detection rather than trajectory forecasting~\cite{dtu_abnormal_ais_dataset}. Piraeus and Brest also provide vessel, geographic, or environmental information, but these data are distributed at the dataset level rather than aligned with fixed forecasting samples.

Several forecasting studies define more detailed experimental settings. TrAISformer~\cite{nguyen2024traisformer} uses a processed Danish AIS subset with a study-specific temporal split and a long-horizon forecasting task, while MSTFormer~\cite{qiang2023mstformer} evaluates multiple forecasting settings constructed from NOAA AIS data. These datasets and protocols are useful for evaluating their associated methods, but remain tied to individual studies. As summarized in Table~\ref{tab:dataset_comparison}, the representative
prior resources either require user-defined task construction or provide study-specific forecasting settings without sample-aligned
external context. None combines fixed multi-horizon tasks,
vessel-disjoint splits, sample-aligned maritime context, and
cross-region evaluation under one common protocol.

\noindent\textbf{AIS-based vessel trajectory forecasting.}
Vessel trajectory forecasting methods range from kinematic and classical machine-learning approaches to recurrent, Transformer-based, and probabilistic models. Recurrent and sequence-to-sequence methods learn temporal motion patterns from historical AIS observations~\cite{capobianco2021deep_vessel,related-lstm-tang2022model,liu2022dl_powered,seq2seq-you2020st}. TrAISformer and MSTFormer use Transformer architectures to capture longer-range temporal dependencies~\cite{nguyen2024traisformer,qiang2023mstformer}. Other studies incorporate geographic constraints, neighboring-vessel interactions, regional motion patterns, or predictive uncertainty~\cite{mehri2021contextual,murray2021regional,rela-GP-rong2019ship,is-gcnn-liang2022fine,gao2024multi,MFPD-ma2025multi,makun10889007}.

Despite these advances, most methods are evaluated on independently processed regional subsets. Differences in data cleaning, trajectory construction, forecasting tasks, and evaluation splits therefore remain difficult to separate from differences in model design. EnvShip is method-agnostic: it fixes these factors and provides a common basis for comparing existing and future forecasting methods under consistent conditions.

\noindent\textbf{Context-aware trajectory benchmarks in neighboring domains.}
Pedestrian forecasting has long relied on shared evaluation resources such as ETH and UCY~\cite{pellegrini2009you,lerner2007crowds} and the Stanford Drone Dataset~\cite{robicquet2016learning}. These datasets have supported the development and comparison of models for social interaction and multimodal motion, including Social-LSTM, Social-GAN, and Social-STGCNN~\cite{alahi2016social,gupta2018socialgan,social-stgcnn-cvpr}. Interaction-rich multi-agent forecasting has also been studied using NBA player-tracking data~\cite{hauri2021nba}.

In autonomous driving, Argoverse and nuScenes combine agent trajectories with map and scene information~\cite{chang2019argoverse,caesar2020nuscenes}, supporting context-aware models such as Trajectron++, Scene Transformer, AgentFormer, and HiVT~\cite{salzmann2020trajectronpp,ngiam2022scene,yuan2021agentformer,zhou2022hivt}. Maritime forecasting has likewise begun to incorporate environmental constraints and vessel interactions~\cite{mehri2021contextual,murray2021regional,is-gcnn-liang2022fine,gao2024multi}, but existing studies still rely largely on study-specific data construction and evaluation.
EnvShip extends the shared-benchmark paradigm to maritime forecasting
by unifying fixed multi-horizon tasks, vessel-disjoint splits,
sample-aligned environmental and neighboring-vessel context, and
cross-region evaluation within a reproducible pipeline. This
controlled design isolates context utility and geographic transfer
from differences in task construction.

\section{The EnvShip Benchmark}
\label{sec:construction}

This section describes the task
definition, shared construction pipeline, track curation, and context
alignment. Complete schemas, directory layouts, and field definitions
are provided with the public dataset release.

\subsection{Data Sources and Forecasting Protocol}
\label{sec:construction_protocol}

EnvShip is constructed from four public AIS sources: the Danish
Maritime Authority archive~\cite{dma_historical_ais}, NOAA
MarineCadastre~\cite{marinecadastre_accessais}, the Piraeus
dataset~\cite{tritsarolis2022piraeus}, and the Norwegian Kystverket
feed~\cite{kystverket}. Table~\ref{tab:source_counts} summarizes the
source periods, anchor extents, unique vessel counts, candidate
windows, and released benchmark samples. Candidate windows are
quality-filtered forecasting windows generated before standard-track
curation. Vessel counts denote unique MMSIs within each source, and
the total is the sum across the four sources.

\begin{table*}[t]
\centering
\caption{Per-region coverage and released benchmark size.
Anchor extents and vessel counts are computed from the released
samples. 
``Filtered'' denotes samples passing the OpenStreetMap
temporal-consistency audit in Section~\ref{sec:construction_qc}}
\label{tab:source_counts}

\footnotesize
\setlength{\tabcolsep}{3.4pt}
\renewcommand{\arraystretch}{1.12}

\resizebox{\textwidth}{!}{%
\begin{tabular}{
p{2.75cm}
p{2.45cm}
p{3.30cm}
r
r
p{3.25cm}
p{3.25cm}
p{3.25cm}
p{3.25cm}}
\toprule

\rowcolor{tableheader}
\textbf{Source / region}
& \textbf{Time span}
& \textbf{Anchor extent}
& \textbf{Vessels}
& \textbf{Candidate windows}
& \multicolumn{2}{c}{\textbf{Track A: 10/10 min}}
& \multicolumn{2}{c}{\textbf{Track B: 30/60 min}} \\

\cmidrule(lr){6-7}
\cmidrule(lr){8-9}

\rowcolor{tablesubheader}
&
&
&
&
&
\textbf{Train / Val / Test}
& \textbf{Full / filtered (ret.)}
& \textbf{Train / Val / Test}
& \textbf{Full / filtered (ret.)} \\

\midrule

DMA / Denmark
& 2025-09-01--2025-09-30
& $52.7$--$59.9^{\circ}$N;
  $2.1$--$19.7^{\circ}$E
& 4{,}014
& 5.48\,M
& 120{,}000 / 15{,}000 / 15{,}000
& 150{,}000 / 148{,}883 (99.26\%)
& 46{,}744 / 5{,}414 / 6{,}000
& 58{,}158 / 57{,}713 (99.24\%) \\

NOAA / United States
& 2025-03-01--2025-03-31
& $13.7$--$49.5^{\circ}$N;
  $159.1^{\circ}$W--$144.3^{\circ}$E
& 3{,}379
& 3.55\,M
& 48{,}000 / 6{,}000 / 6{,}000
& 60{,}000 / 59{,}580 (99.30\%)
& 35{,}893 / 4{,}725 / 4{,}148
& 44{,}766 / 37{,}612 (84.02\%) \\

Piraeus / Greece
& 2019-01-01--2019-12-26
& $37.7$--$38.0^{\circ}$N;
  $23.1$--$23.7^{\circ}$E
& 1{,}775
& 158\,K
& 48{,}000 / 6{,}000 / 6{,}000
& 60{,}000 / 59{,}770 (99.62\%)
& 628 / 394 / 132
& 1{,}154 / 1{,}154 (100.00\%) \\

Kystverket / Norway
& 2025-08-01--2025-09-30
& $57.5$--$60.5^{\circ}$N;
  $4.0$--$11.9^{\circ}$E
& 687
& 289\,K
& 48{,}000 / 6{,}000 / 6{,}000
& 60{,}000 / 58{,}083 (96.81\%)
& 2{,}441 / 117 / 221
& 2{,}779 / 2{,}712 (97.59\%) \\

\midrule

\rowcolor{tabletotal}
\textbf{Total (4 regions)}
& ---
& ---
& \textbf{9{,}855}
& \textbf{9.48\,M}
& \textbf{264{,}000 / 33{,}000 / 33{,}000}
& \textbf{330{,}000 / 326{,}316 (98.88\%)}
& \textbf{85{,}706 / 10{,}650 / 10{,}501}
& \textbf{106{,}857 / 99{,}191 (92.83\%)} \\

\bottomrule
\end{tabular}%
}
\end{table*}

Track~A and Track~B are generated from the same regional trajectory
cores but impose different duration requirements and forecasting
windows.
All trajectories are sampled at 20-second intervals and represented
in a target-vessel-centered metric frame. Geographic coordinates are
first projected into a local east--north coordinate system. The final
observed position, referred to as the \emph{anchor}, is translated to
the origin, and the coordinate frame is rotated according to the final
observed heading.
Track~A contains $T_h=30$ observed and $T_f=30$ future positions,
corresponding to 10 minutes of observation and 10 minutes of
prediction. Track~B contains $T_h=90$ observed and $T_f=180$ future
positions, corresponding to 30 minutes of observation and 60 minutes
of prediction. Let

\[
\mathbf{X}=\{(x_t,y_t)\}_{t=1}^{T_h},
\qquad
\mathbf{Y}=\{(x_t,y_t)\}_{t=T_h+1}^{T_h+T_f}.
\]

The task is to predict the future trajectory $\mathbf{Y}$ from the
observed $\mathbf{X}$, optionally conditioned on the
sample-aligned contextual data.

\subsection{Shared Construction Pipeline}
\label{sec:construction_core}

Only the source adapter is source-specific, and all downstream stages
follow the same processing pipeline, with shared parameters defined
in version-controlled configuration files.

\noindent\textbf{Normalization and filtering.}
Records are mapped to a common schema containing MMSI, UTC timestamp,
position, speed over ground (SOG), course over ground (COG), heading,
navigation status, and available static vessel attributes.
Source-specific vessel types are mapped to a shared eight-category
taxonomy.
Records with invalid identifiers, timestamps, positions, or speeds are
removed, and source-specific AIS sentinel values are treated as
missing. Duplicate vessel--timestamp reports are resolved by retaining
the record with the most complete fields. Motion screening applies a
global speed limit of 55\,kn and category-specific limits of 35\,kn
for cargo vessels and tankers, 45\,kn for passenger vessels and
ferries, 25\,kn for fishing vessels, and 20\,kn for tug and service
vessels.

\noindent\textbf{Segmentation and resampling.}
Records are ordered by vessel and divided at temporal gaps exceeding
600\,s, non-monotonic timestamps, or physically implausible implied
motion. Segments dominated by anchored, moored, or near-stationary
behavior are excluded.
Gaps of up to 120\,s are interpolated in a local metric frame, with
circular interpolation applied to COG and heading. Longer gaps are
not bridged. Valid segments are resampled on a common 20-second UTC
grid, screened again for geometric inconsistencies, and required to
contain at least 80 points.

\noindent\textbf{Window construction and quality metadata.}
Sliding windows are extracted under the two forecasting protocols.
Each sample is accompanied by quality metadata describing its
interpolation ratio, displacement, path efficiency, speed variation,
and temporal continuity, together with an environmental scene label.
Motivated by the challenging-scenario curation of Argoverse~2
~\cite{wilson2021argoverse2} and the scenario-stratified evaluation
of CRITERIA~\cite{chen2024criteria}, we introduce a maritime-specific
trajectory difficulty score based on the mean and peak turning behavior
over the complete observed-and-future window. Larger and more frequent
turns indicate greater forecasting difficulty. The score is used only
for sample curation and stratified evaluation, not as a model input:

\begin{equation}
D_i =
\bar{\theta}^{\mathrm{hist}}_i
+
\bar{\theta}^{\mathrm{fut}}_i
+
0.5\!\left(
\theta^{\max,\mathrm{hist}}_i
+
\theta^{\max,\mathrm{fut}}_i
\right)
+
\theta^{\mathrm{bridge}}_i
+
e^{\mathrm{cog}}_i,
\label{eq:difficulty}
\end{equation}

where $\bar{\theta}$ and $\theta^{\max}$ denote the mean and maximum
absolute stepwise turning angles over the observed or future
trajectory. The term $\theta^{\mathrm{bridge}}$ measures the turn
across the observation--prediction boundary, and
$e^{\mathrm{cog}}$ is the mean absolute difference between reported
COG and motion-derived heading. All angular differences are circularly
wrapped and expressed in degrees.
Samples are assigned to easy ($D_i<5$), medium
($5\leq D_i<12$), and hard ($D_i\geq12$) tiers. Because this score
uses the future trajectory, it is used only for sample curation and
stratified evaluation, not as a forecasting input.

Samples are assigned to the training, validation, and test sets at an
80/10/10 ratio using a deterministic hash of the MMSI. The resulting
splits are vessel-disjoint, preventing windows from the same vessel
from appearing in multiple splits.



\begin{figure*}[t]
\centering
\includegraphics[
    width=0.94\textwidth,
    trim=10 3 6 4,
    clip
]{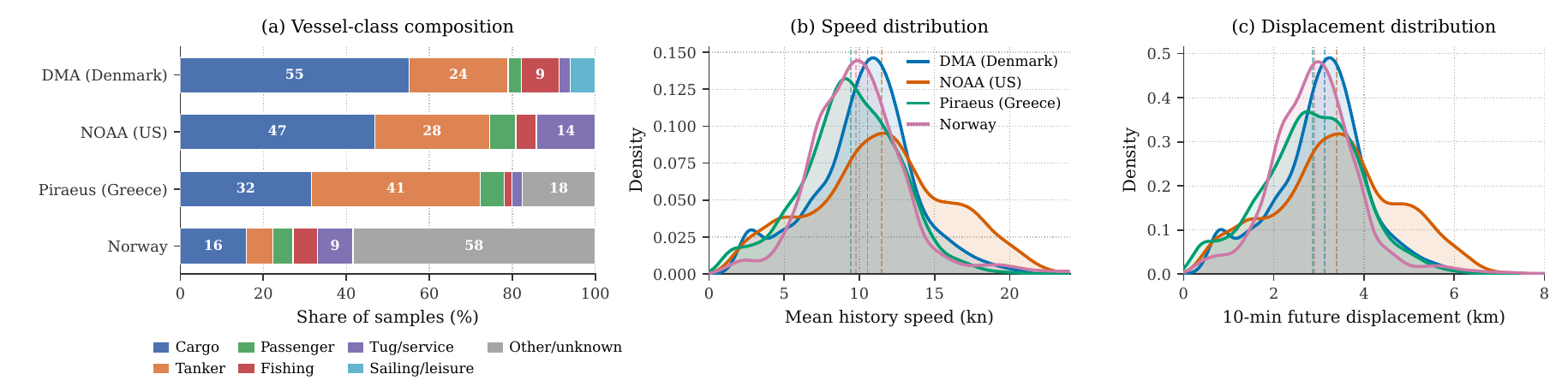}
\caption{Regional statistics for Track~A (330{,}000 samples). Dashed lines in (b) and (c)
denote regional medians.}
\label{fig:benchmark_basic_stats}
\end{figure*}
\subsection{Standard-Track Curation}
\label{sec:construction_lite}

The standard tracks are curated from the regional core sets through
vessel-class stratification, difficulty balancing, quality-based
ranking, and redundancy control. Vessel-class proportions are controlled
at the benchmark level rather than enforced independently for each
region, allowing the curated tracks to retain regional differences in
fleet composition and operating patterns. Samples are balanced across
easy, medium, and hard cases at a ratio of 35/40/25. As shown in
Fig.~\ref{fig:benchmark_basic_stats}, the four regions remain distinct
in their vessel-class composition, historical speed distribution, and
10-minute future displacement.
To reduce near-duplicate motion windows, selections are capped per
vessel and continuous segment, with a minimum temporal separation
between windows from the same segment. The training split contains at
most 55 windows per vessel and 12 per segment; for validation and test,
the corresponding caps are 70 and 20. Track~A therefore contains
330{,}000 samples: 150{,}000 from DMA and 60{,}000 from each of the
other three sources.

\begin{figure}[t]
    \centering
    \includegraphics[width=0.98\columnwidth]{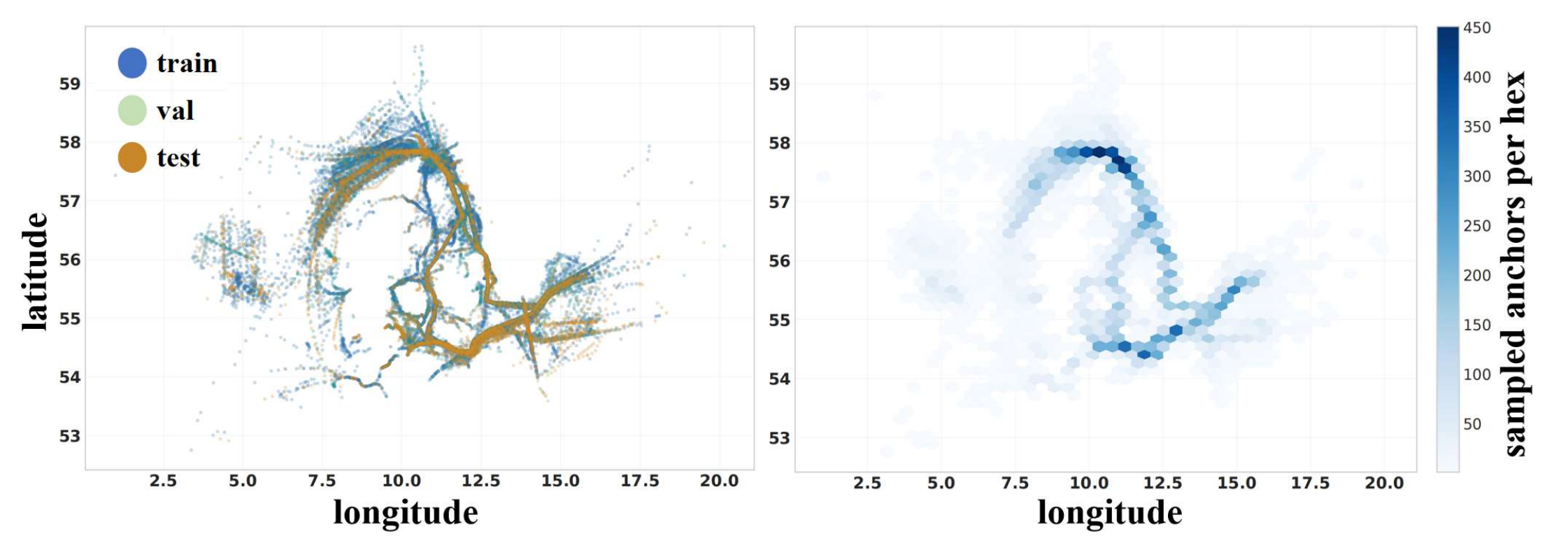}
    \caption{\textbf{Geographic coverage of DMA Track~A anchors.}
    Left: anchor locations in the training, validation, and test
    splits. Right: hexagonal-bin density of all released anchors.
    }
    \label{fig:geo_coverage}
\end{figure}

Figure~\ref{fig:geo_coverage} illustrates the geographic coverage
of the DMA Track~A anchors. The training, validation, and test
splits broadly cover the same principal operating areas, while the
density map highlights the concentration of samples along major
traffic corridors and other high-traffic areas. Because split
assignment is vessel-disjoint, this geographic overlap does not
introduce vessel-level leakage.

\begin{figure}[t]
    \centering
    \includegraphics[width=0.98\columnwidth]{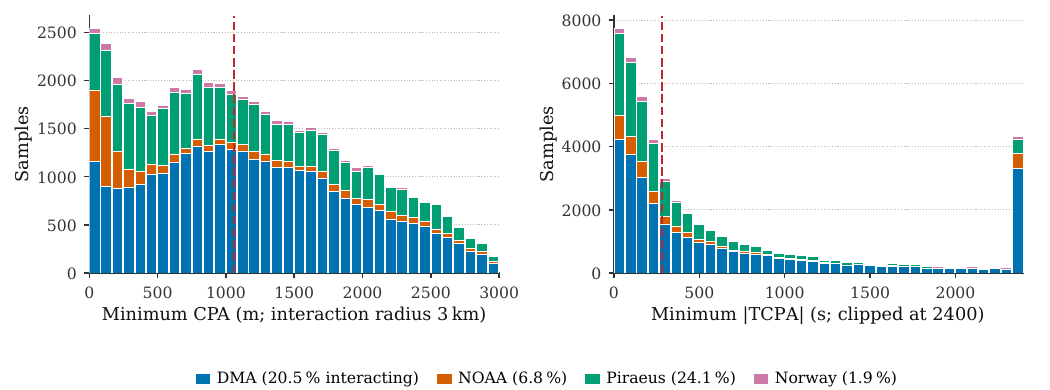}
    \caption{\textbf{Neighboring-vessel interaction statistics for
    Track~A.} Regional distributions of minimum CPA and absolute TCPA;
    legend indicates the proportion of interacting samples.}
    \label{fig:cpa_tcpa}
\end{figure}

\subsection{Context Packages}
\label{sec:construction_context}

A stable sample identifier links each forecasting window to its
context records while preserving the trajectory target and split
assignment.

\noindent\textbf{Environmental context.}
For both tracks, OpenStreetMap coastlines, waterways, piers,
breakwaters, quays, and port-related features are projected into the
target-centered local frame and cropped once at the anchor to an
axis-aligned $10\,\mathrm{km}\times10\,\mathrm{km}$ window
($\pm5$\,km per axis). The package includes a six-channel
$128\times128$ raster encoding land, water, navigable water, natural
and man-made boundaries, and barriers; two signed-distance fields,
$d_{\mathrm{shore}}$ and $d_{\mathrm{nav}}$; clipped vector
polylines; and a 14-dimensional scene descriptor summarizing scene
type, distance, area ratio, boundary density, and context quality.

\noindent\textbf{Neighboring-vessel context.}
For each target sample, up to ten nearest vessels within a 3\,km
radius are selected according to their distance at the anchor time.
The released context records contain target-relative neighbor
trajectories, relative velocities, pairwise distances, and validity
masks that indicate the availability of neighboring-vessel states.
They also include the Closest Point of Approach (CPA) and Time to
Closest Point of Approach (TCPA) for each target--neighbor pair.
Both quantities are computed exclusively from motion states
available at the anchor time and therefore do not use any information
from the prediction interval. Figure~\ref{fig:cpa_tcpa} reports the
regional distributions of the minimum CPA and absolute TCPA in
Track~A, together with the proportion of samples containing at least
one neighboring-vessel interaction.

\noindent\textbf{Weather, sea-state, and traffic attributes.}
Each sample includes six weather variables from the ERA5 reanalysis
dataset, four wave and sea-state variables from the European Centre for
Medium-Range Weather Forecasts (ECMWF) Wave Model, information about
the nearest port, and three indicators describing nearby fairways and
traffic-separation schemes derived from OpenStreetMap. If a data source
does not cover a particular sample, the corresponding attributes are
marked as unavailable rather than estimated or imputed.

\subsection{OpenStreetMap Temporal-Consistency Audit}
\label{sec:construction_qc}

OpenStreetMap~\cite{haklay2008openstreetmap,openstreetmap} changes over time. Coastal infrastructure added after
the AIS observation period, including reclaimed land, piers, and
breakwaters, may cause historically valid vessel trajectories to
appear inland when evaluated against a later map snapshot.

The audit projects each trajectory point onto its shoreline
signed-distance field and records maximum inland penetration, the
number of inland points, and the longest consecutive inland sequence.
A sample passes the recommended filter when its inland depth does not
exceed 30\,m and fewer than three consecutive points are inland.

No sample is deleted. The audit flags and diagnostic values are
included in the release, allowing users to select either the complete
or recommended subset. The filter retains 326{,}316 of the 330{,}000
Track~A samples and 99{,}191 of the 106{,}857 Track~B samples.

Because the Piraeus AIS observations were collected in 2019, its
environmental context is additionally constructed using a pinned
OpenStreetMap snapshot dated 2020-01-01. The corresponding consistency
audit uses the same snapshot. Unless stated otherwise, all experiments
use the recommended temporally consistent subset.


\begin{figure*}[t]
    \centering
    \includegraphics[width=0.8\textwidth]{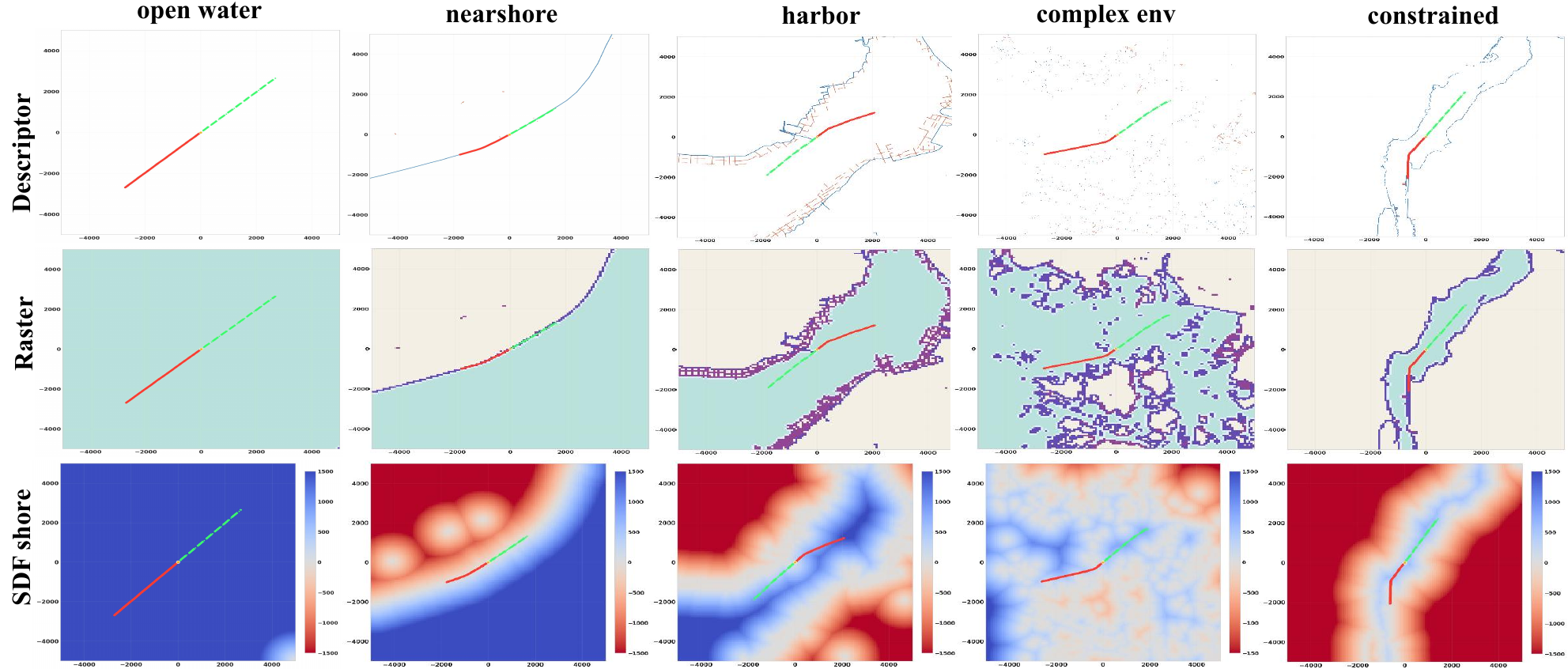}
    \caption{
    Representative environmental-context samples.
Columns show different maritime scenes; rows show descriptor(vector geometry),
water--land rasters, and shoreline signed-distance fields.}
    \label{fig:context_cases}
\end{figure*}

\section{Evaluation Setup}
\label{sec:analysis}

\subsection{Tasks and Metrics}
\label{sec:analysis_eval}

EnvShip evaluates physics-based, classical machine-learning,
trajectory-only deep, and context-aware models under a common
forecasting protocol. The learned context-aware comparison uses four
matched input configurations: trajectory-only, neighboring-vessel-aware,
environment-aware, and joint-context forecasting. Prediction targets,
vessel-disjoint splits, coordinate frames, and evaluation metrics are
held fixed across models. Hyperparameters and checkpoints are selected
on the validation split, and final results are reported on the held-out
test split.

The primary metrics are average displacement error (ADE) and final
displacement error (FDE), measured in meters:

\begin{equation}
\mathrm{ADE}=\frac{1}{NT}\sum_{i=1}^{N}\sum_{t=1}^{T}
\left\|\hat{\mathbf{y}}_{i,t}-\mathbf{y}_{i,t}\right\|_2,
\qquad
\mathrm{FDE}=\frac{1}{N}\sum_{i=1}^{N}
\left\|\hat{\mathbf{y}}_{i,T}-\mathbf{y}_{i,T}\right\|_2.
\label{eq:ade_fde}
\end{equation}
where $N$ is the number of test samples and $T=T_f$ is the
prediction length. Predictions and targets are denormalized to the
target-vessel-centered metric frame before evaluation.

Beyond full-horizon ADE and FDE, we evaluate both tracks at
multiple intermediate prediction horizons. Owing to space
constraints, the main paper reports only representative checkpoints,
including 3- and 6-minute ADE for Track~A and selected long-horizon
results for Track~B. Complete per-horizon results for all evaluated
models and regions are provided in the public benchmark release.
We further stratify ADE by trajectory difficulty, scene type, and
neighboring-vessel count. For methods that generate multiple future
trajectories, the evaluation toolkit additionally supports
$\mathrm{minADE}@K$ and $\mathrm{minFDE}@K$, computed as the
minimum error over the $K$ predicted futures for each sample.

\subsection{Reference Models}
\label{sec:analysis_baselines}

The main comparison contains multiple reference systems selected to cover
complementary motion assumptions, temporal architectures, and context
representations.

\noindent\textbf{Physics and classical machine learning.}
Constant Velocity extrapolates the mean velocity over the final three
observed steps, whereas Dead Reckoning integrates the final reported
speed over ground and course over ground. Random Forest
~\cite{breiman2001random} predicts each future position from flattened
trajectory and motion features using a separate two-dimensional
regressor for each prediction step.

\noindent\textbf{Trajectory-only models.}
We include a two-layer LSTM encoder--decoder with 256 hidden units
~\cite{hochreiter1997long}, a Temporal Convolutional Network with
dilated causal convolutions~\cite{bai2018empirical}, and a
non-autoregressive Transformer that predicts the complete future
sequence in one forward pass
~\cite{vaswani2017attention,giuliari2021transformer}. These models
represent recurrent, convolutional, and attention-based temporal
architectures, respectively.

\noindent\textbf{Neighboring-vessel context.}
Both neighboring-vessel models use the same LSTM trajectory encoder
and a shared MLP to embed each valid neighbor. The pooling model
aggregates these embeddings into a single context vector. The gated
attention model instead performs validity-masked cross-attention at
each decoder step, incorporates CPA, TCPA, and relative-speed
attributes, and uses a neighbor-count gate to suppress the interaction
pathway when no valid neighbors are present.

\noindent\textbf{Environmental context.}
The descriptor model conditions the decoder at each prediction step
through feature-wise linear modulation (FiLM)
~\cite{perez2018film}. Raster-based models form a controlled
$2\times2$ comparison of
$\{\text{binary raster},\,\text{SDF}\}\times
\{\text{global pooling},\,\text{spatial attention}\}$.
The binary raster provides discrete occupancy information, whereas the
signed-distance fields(SDF) encode continuous distances to shoreline and
navigable-region boundaries.
Representative
environmental representations across different maritime scenes are
shown in Figure~\ref{fig:context_cases}.
Global-pooling models compress the CNN
feature map into one vector; spatial-attention models retain the
feature map and query it at each decoder step. This design separates
the effect of environmental representation from that of context
aggregation.

\noindent\textbf{Joint context.}
The descriptor-based joint model combines neighboring-vessel
attention with FiLM conditioning from the environmental descriptor.
The SDF-based joint model combines gated neighboring-vessel attention
with spatial attention over the SDF feature map. These models evaluate
whether interaction and geographic information provide complementary
forecasting cues.

\subsection{Training and Statistical Protocol}
\label{sec:analysis_protocol}

Neural models are optimized with AdamW
~\cite{loshchilov2019decoupled} using a learning rate of
$5\times10^{-4}$, weight decay of $10^{-4}$, a cosine
learning-rate schedule, Huber loss with $\delta=1$ in normalized
coordinate space~\cite{huber1964robust}, and a batch size of 256.
Training runs for at most 80 epochs and stops after 20 epochs
without improvement in validation loss. For autoregressive recurrent
decoders, teacher forcing is linearly annealed to zero over the first
40\% of training.
For each training configuration, normalization statistics are
estimated exclusively from its training split and applied unchanged
to the corresponding validation and test samples. Checkpoints are
selected by validation loss. Unless a multimodal head is explicitly
enabled, each learned model outputs one deterministic future
trajectory.



\newcommand{\pmstd}[1]{\,\raisebox{0.15ex}{\tiny$\pm #1$}}

\begin{table*}[t]
\centering
\caption{DMA Track~A in-domain results. For learned models, ADE and
FDE are reported as mean $\pm$ standard deviation across 5-seeds;
temporal, difficulty, and scene entries report seed-averaged ADE.
All values are in meters and
lower is better. The best mean in each column is shown in bold.}
\label{tab:dma_main}

\scriptsize
\setlength{\tabcolsep}{1.75pt}
\renewcommand{\arraystretch}{1.02}

\resizebox{\textwidth}{!}{%
\begin{tabular}{
>{\centering\arraybackslash}p{1.50cm}
>{\raggedright\arraybackslash}p{3.45cm}
c c
!{\vrule width 0.35pt}
c c
!{\vrule width 0.35pt}
c c c
!{\vrule width 0.35pt}
c c c}
\toprule

\rowcolor{black!6}
\multicolumn{1}{c}{\rule{0pt}{2.5ex}}
& \multicolumn{1}{c}{}
& \multicolumn{1}{c}{}
& \multicolumn{1}{c!{\vrule width 0.35pt}}{}
& \multicolumn{2}{c!{\vrule width 0.35pt}}{\textbf{Temporal}}
& \multicolumn{3}{c!{\vrule width 0.35pt}}{\textbf{Difficulty}}
& \multicolumn{3}{c}{\textbf{Scene}} \\

\rowcolor{black!6}
\rule{0pt}{2.5ex}\textbf{Category}
& \textbf{Model}
& \textbf{ADE}$\downarrow$
& \textbf{FDE}$\downarrow$
& \textbf{3 min}
& \textbf{6 min}
& \textbf{Easy}
& \textbf{Medium}
& \textbf{Hard}
& \textbf{Open}
& \textbf{Nearshore}
& \textbf{Harbor} \\
\midrule

\multirow{2}{*}{\textbf{Physics}}
& Constant Velocity
& 98.4
& 243.6
& 18.8
& 45.9
& 29.0
& 107.8
& 404.8
& 94.5
& 187.0
& 201.4 \\

& Dead Reckoning
& 91.6
& 231.9
& 16.0
& 41.3
& \textbf{26.5}
& 98.5
& 382.5
& 88.0
& 171.9
& 174.7 \\
\midrule

\textbf{Classical ML}
& Random Forest
& 96.1
& 245.0
& 16.7
& 43.0
& 43.7
& 100.9
& 332.4
& 93.5
& 149.1
& 148.8 \\
\midrule

\multirow{3}{*}{\textbf{\shortstack{Trajectory\\only}}}
& LSTM
& 93.2\pmstd{0.5}
& 236.0\pmstd{2.1}
& 18.9
& 42.8
& 42.7
& 98.6
& 319.3
& 93.2
& 167.9
& 154.9 \\

& TCN
& 87.0\pmstd{0.7}
& 222.0\pmstd{1.4}
& \textbf{15.9}
& \textbf{40.5}
& 36.8
& \textbf{94.4}
& 307.7
& 87.0
& 145.2
& 140.7\\

& Transformer-NAR
& 95.7\pmstd{1.6}
& 234.7\pmstd{3.5}
& 23.1
& 46.4
& 46.8
& 101.7
& 313.4
& 95.7
& 149.7
& 158.8 \\
\midrule

\multirow{2}{*}{\textbf{\shortstack{Neighbor\\context}}}
& LSTM + Neighbor Pool
& 91.1\pmstd{1.1}
& 229.8\pmstd{2.4}
& 18.6
& 42.1
& 40.2
& 98.3
& 315.5
& 87.3
& 168.0
& 147.9 \\

& LSTM + Gated Neighbor Attention
& 94.3\pmstd{0.2}
& 238.3\pmstd{1.4}
& 18.9
& 43.6
& 44.2
& 99.0
& 319.8
& 90.6
& 168.5
& 155.5 \\
\midrule

\multirow{5}{*}{\textbf{\shortstack{Environment\\context}}}
& LSTM + Descriptor FiLM
& 87.0\pmstd{0.2}
& 213.0\pmstd{0.9}
& 21.0
& 42.3
& 39.0
& 98.4
& 290.2
& 83.5
& 155.2
& 148.4 \\

& LSTM + Binary Raster
& 86.2\pmstd{0.4}
& 211.2\pmstd{1.0}
& 18.7
& 40.9
& 38.8
& 97.8
& 279.5
& \textbf{82.2}
& \textbf{144.9}
& 140.7 \\

& LSTM + Binary Spatial Attention
& 89.0\pmstd{1.2}
& 219.6\pmstd{1.2}
& 18.4
& 42.4
& 42.3
& 99.1
& 288.1
& 85.9
& 156.0
& 145.7 \\

& LSTM + SDF
& \textbf{85.5}\pmstd{0.9}
& \textbf{210.3}\pmstd{2.4}
& 17.8
& 41.0
& 38.9
& 96.6
& 282.2
& 82.3
& 157.9
& \textbf{134.6} \\

& LSTM + SDF Spatial Attention
& 87.2\pmstd{0.2}
& 215.0\pmstd{1.2}
& 18.7
& 42.0
& 41.5
& 97.5
& 281.9
& 84.3
& 149.6
& 149.4 \\
\midrule

\multirow{2}{*}{\textbf{\shortstack{Joint\\context}}}
& LSTM + Neighbor Attn + Descriptor
& 87.0\pmstd{0.4}
& 213.4\pmstd{0.6}
& 20.3
& 42.0
& 39.5
& 97.9
& 288.7
& 83.5
& 152.9
& 159.6 \\

& LSTM + Gated Neighbor Attn + SDF
& 87.4\pmstd{0.4}
& 215.1\pmstd{2.1}
& 18.6
& 42.0
& 42.9
& 97.9
& \textbf{275.8}
& 84.6
& 148.5
& 141.2 \\
\bottomrule
\end{tabular}%
}

\end{table*}

\section{In-Domain Results}
\label{sec:indomain}

We summarize representative in-domain results for DMA Track~A,
regional variation on Track~A, and the long-horizon Track~B.
Models are trained and selected on the corresponding regional
training and validation sets and evaluated on the vessel-disjoint
test set using the recommended temporally consistent subset.
Complete model--region and intermediate-horizon results are
provided in the benchmark release.

\subsection{Short-Horizon Track~A}
\label{sec:indomain_dma}

Dead Reckoning remains a strong reference at 91.6\,m ADE,
outperforming Random Forest and the plain LSTM. TCN is the
strongest trajectory-only model, achieving
$87.0\pm0.7$\,m ADE, as shown in Table~\ref{tab:dma_main}. Environment-aware models obtain the lowest
aggregate errors: LSTM+SDF achieves
$85.5\pm0.9$\,m ADE and $210.3\pm2.4$\,m FDE, followed closely
by Binary Raster at $86.2\pm0.4$\,m ADE.
Spatial attention does not provide a consistent improvement.
It increases ADE from 86.2 to 89.0\,m for binary rasters and from
85.5 to 87.2\,m for SDFs. The results therefore support the utility
of environmental conditioning, but do not establish a uniform
advantage for either spatial attention or SDF representation.
The preferred representation also varies by scene. Binary Raster
achieves the lowest nearshore ADE at 144.9\,m, whereas SDF performs
best in harbor scenes at 134.6\,m. Because the harbor cohort contains
only 15 samples, this comparison is diagnostic rather than conclusive.
Even the best hard-tier result remains 275.8\,m ADE, confirming the
difficulty of strongly turning trajectories.

\noindent\textbf{Neighboring-vessel context.}
\label{sec:social}
Neighbor Pool modestly improves over the LSTM
($91.1$ versus $93.2$\,m ADE), whereas Gated Neighbor Attention
does not ($94.3$\,m). However, 79.5\% of the DMA test samples contain
no neighboring vessel within 3\,km, so aggregate results are dominated
by isolated trajectories. Neighbor-count-stratified results are provided in the public
benchmark release.

\noindent\textbf{Seed stability.}
\label{sec:multiseed}
Seed-level variation is generally modest, with ADE standard
deviations between 0.2 and 1.6\,m across learned models.
The differences among SDF ($85.5\pm0.9$\,m), Binary Raster
($86.2\pm0.4$\,m), TCN ($87.0\pm0.7$\,m), and Descriptor FiLM
($87.0\pm0.2$\,m) are therefore small relative to their variability.
Close rankings should consequently be interpreted as comparable
performance rather than robust superiority.

\paragraph{Regional Variation on Track~A}
\label{sec:indomain_xjuris}

Model rankings vary substantially across operating regimes. On NOAA,
TCN performs best at 92.8\,m ADE, while SDF reaches 100.4\,m,
indicating limited aggregate benefit from environmental context in
the predominantly open-water setting. On Norway, Dead Reckoning
achieves 108.4\,m ADE and outperforms the best learned model
(SDF, 127.5\,m). In contrast, SDF achieves 151.8\,m on Piraeus,
improving over TCN at 169.1\,m by 17.3\,m. Environmental context
therefore provides its clearest benefit where geographic constraints
are informative, rather than uniformly across regions.

\subsection{Long-Horizon Track~B}
\label{sec:trackb}

\providecommand{\pmstd}[1]{\,{\tiny$\pm#1$}}

\begin{table}[t]
\centering
\caption{Track~B full-horizon in-domain results
($30\,\mathrm{min}\!\rightarrow\!60\,\mathrm{min}$) on DMA
($N{=}6{,}000$) and NOAA ($N{=}4{,}148$). For ADE and FDE (m), lower is
better. Best values are shown in bold.}
\label{tab:trackb}

\scriptsize
\setlength{\tabcolsep}{2.6pt}
\renewcommand{\arraystretch}{1.02}

\resizebox{\columnwidth}{!}{%
\begin{tabular}{l cc cc}
\toprule
& \multicolumn{2}{c}{\textbf{DMA Track~B}}
& \multicolumn{2}{c}{\textbf{NOAA Track~B}} \\
\cmidrule(lr){2-3}
\cmidrule(lr){4-5}
\textbf{Model}
& \textbf{ADE}$\downarrow$
& \textbf{FDE}$\downarrow$
& \textbf{ADE}$\downarrow$
& \textbf{FDE}$\downarrow$ \\
\midrule

LSTM
& 658.5\pmstd{2.2}
& 1539.2\pmstd{9.5}
& 1361.1\pmstd{5.3}
& 3055.3\pmstd{15.1} \\

GRU
& 642.6\pmstd{3.8}
& \textbf{1486.8}\pmstd{7.6}
& 1208.1\pmstd{6.2}
& 2693.8\pmstd{13.5} \\

BiLSTM
& 651.5\pmstd{2.1}
& 1536.7\pmstd{7.3}
& 1358.9\pmstd{4.8}
& 3022.6\pmstd{16.3} \\

TCN
& \textbf{635.1}\pmstd{2.3}
& 1489.6\pmstd{5.8}
& \textbf{1136.5}\pmstd{4.4}
& \textbf{2575.2}\pmstd{10.2} \\

Transformer-NAR
& 680.4\pmstd{1.9}
& 1545.8\pmstd{16.7}
& 1247.9\pmstd{3.8}
& 2750.3\pmstd{28.4} \\

LSTM + Neighbor Pool
& 652.0\pmstd{3.4}
& 1543.8\pmstd{12.4}
& 1395.1\pmstd{6.9}
& 3146.8\pmstd{32.1} \\

LSTM + Binary Raster
& 643.1\pmstd{2.9}
& 1510.6\pmstd{10.1}
& 1872.7\pmstd{7.9}
& 4189.1\pmstd{30.5} \\

LSTM + Binary SpAttn
& 659.3\pmstd{4.8}
& 1544.4\pmstd{6.2}
& 1232.8\pmstd{7.4}
& 2730.7\pmstd{23.1} \\

LSTM + SDF
& 643.7\pmstd{2.2}
& 1521.5\pmstd{10.6}
& 1369.4\pmstd{5.8}
& 3085.9\pmstd{22.7} \\

LSTM + G-Nbr Attn + SDF
& 663.1\pmstd{4.4}
& 1551.1\pmstd{15.1}
& 1299.5\pmstd{8.3}
& 2887.7\pmstd{32.6} \\
\bottomrule
\end{tabular}%
}
\end{table}
Track~B extends prediction to 60 minutes. Table~\ref{tab:trackb}
reports full-horizon results for DMA and NOAA, the two largest
Track~B test cohorts; complete four-region and intermediate-horizon
results are provided in the benchmark release.

TCN obtains the lowest ADE on both DMA and NOAA, reaching
635.1 and 1{,}136.5\,m, respectively. GRU achieves the lowest DMA
FDE at 1{,}486.8\,m, narrowly outperforming TCN at 1{,}489.6\,m,
whereas TCN remains best on NOAA for both metrics. Environmental
models remain close to TCN on DMA but degrade more substantially
on NOAA, confirming that model rankings depend on both horizon and
regional operating conditions.

On DMA, SDF reduces ADE relative to the same-backbone LSTM from
658.5 to 643.7\,m, corresponding to a 2.2\% improvement at
60 minutes, compared with approximately 10\% over the first
10 minutes. This attenuation is consistent with the fixed,
anchor-centered context: its contribution decreases as vessels move
beyond the spatial support available at the observation endpoint.

\section{Cross-Region Transfer}
\label{sec:cross_domain}

All regions share the same construction pipeline, forecasting
protocols, and target-centered coordinate frame. Normalization
statistics are estimated from each training pool and applied
unchanged at test time, isolating geographic and operational
distribution shift. We examine Track~A zero-shot transfer and
multi-source source-removal ablations; complete results for both
tracks are provided in the benchmark release.

\subsection{DMA-to-NOAA Zero-Shot Transfer}
\label{sec:zeroshot}

Without NOAA fine-tuning, ADE increases from 87.0 to 101.5\,m for
TCN, from 87.0 to 107.4\,m for Descriptor FiLM, and from 85.5 to
105.5\,m for SDF. Dead Reckoning similarly increases from 91.6 to
106.5\,m, confirming a clear regional shift under otherwise identical
evaluation protocols.
TCN achieves the lowest aggregate NOAA ADE. On nearshore samples,
however, SDF performs best at 203.0\,m, compared with 222.1\,m for
TCN and 251.5\,m for Dead Reckoning. Environmental context therefore
remains useful under geographic shift when coastline geometry
constrains vessel motion.

\subsection{Multi-Source Training and Source Removal}
\label{sec:xdomain_multi}

\begin{figure}[t]
    \centering
    \includegraphics[width=0.96\columnwidth]
    {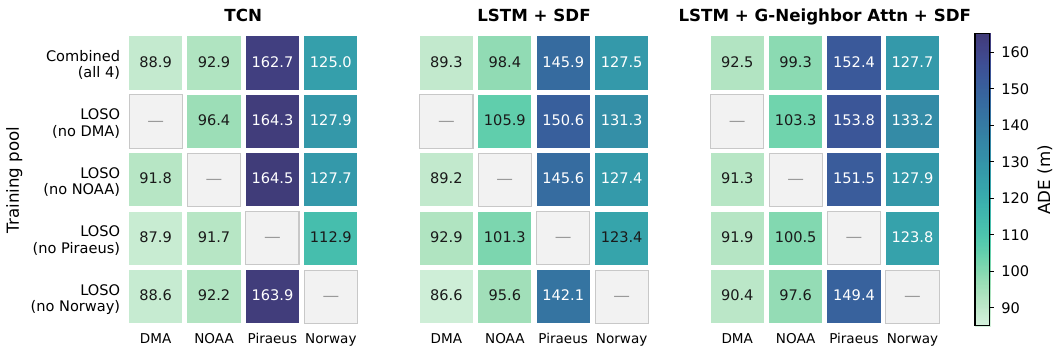}
    \caption{Track~A ADE under multi-source and source-removal training.
Blank cells denote excluded sources.}
    \label{fig:xdomain_heatmap}
\end{figure}

Table~\ref{tab:xdomain_a} and Figure~\ref{fig:xdomain_heatmap}
compare training on all four regions with source-removal variants.
Each removed source is excluded from both training and evaluation;
the remaining cells measure its contribution to the retained regions.

Combined training is competitive but not uniformly beneficial.
TCN achieves 88.9 and 92.9\,m ADE on DMA and NOAA, close to its
in-domain results of 87.0 and 92.8\,m. On Piraeus, combined training
improves TCN from 169.1 to 162.7\,m and SDF from 151.8 to
145.9\,m. Combined SDF also improves NOAA from 100.4 to 98.4\,m,
but degrades DMA from 85.5 to 89.3\,m.

Source contributions are strongly asymmetric. Removing DMA degrades
all three models on most retained regions, whereas removing Norway
improves both context-aware models on every retained target.
Removing Piraeus also reduces Norway ADE from 125.0 to 112.9\,m
for TCN and from 127.5 to 123.4\,m for SDF. Thus, additional
regional data can introduce negative transfer, and effective source
selection depends on both the target regime and model architecture.
The same qualitative behavior is also observed on Track~B.

\begin{table}[t]
\centering
\caption{
Track~A ADE (m) under multi-source and source-removal
training. Values are mean $\pm$ standard deviation over five seeds;
dashes denote excluded sources.}
\label{tab:xdomain_a}

\scriptsize
\setlength{\tabcolsep}{2.8pt}
\renewcommand{\arraystretch}{0.98}

\resizebox{\columnwidth}{!}{%
\begin{tabular}{llcccc}
\toprule
\textbf{Training pool}
& \textbf{Model}
& \textbf{DMA}
& \textbf{NOAA}
& \textbf{Piraeus}
& \textbf{Norway} \\
\midrule

\multirow{3}{*}{Combined (all 4)}
& TCN
& 88.9\pmstd{0.7}
& 92.9\pmstd{0.8}
& 162.7\pmstd{0.8}
& 125.0\pmstd{0.9} \\

& LSTM + SDF
& 89.3\pmstd{1.4}
& 98.4\pmstd{0.9}
& 145.9\pmstd{1.6}
& 127.5\pmstd{1.4} \\

& LSTM + G-Nbr Attn + SDF
& 92.5\pmstd{0.7}
& 99.3\pmstd{1.0}
& 152.4\pmstd{1.3}
& 127.7\pmstd{0.8} \\
\midrule

\multirow{3}{*}{LOSO (no DMA)}
& TCN
& ---
& 96.4\pmstd{1.6}
& 164.3\pmstd{0.5}
& 127.9\pmstd{2.3} \\

& LSTM + SDF
& ---
& 105.9\pmstd{1.9}
& 150.6\pmstd{1.2}
& 131.3\pmstd{2.5} \\

& LSTM + G-Nbr Attn + SDF
& ---
& 103.3\pmstd{1.4}
& 153.8\pmstd{0.7}
& 133.2\pmstd{0.4} \\
\midrule

\multirow{3}{*}{LOSO (no NOAA)}
& TCN
& 91.8\pmstd{1.7}
& ---
& 164.5\pmstd{1.5}
& 127.7\pmstd{2.6} \\

& LSTM + SDF
& 89.2\pmstd{1.5}
& ---
& 145.6\pmstd{1.6}
& 127.4\pmstd{2.5} \\

& LSTM + G-Nbr Attn + SDF
& 91.3\pmstd{0.6}
& ---
& 151.5\pmstd{1.1}
& 127.9\pmstd{1.0} \\
\midrule

\multirow{3}{*}{LOSO (no Piraeus)}
& TCN
& 87.9\pmstd{0.4}
& 91.7\pmstd{0.4}
& ---
& 112.9\pmstd{0.7} \\

& LSTM + SDF
& 92.9\pmstd{1.3}
& 101.3\pmstd{1.4}
& ---
& 123.4\pmstd{2.9} \\

& LSTM + G-Nbr Attn + SDF
& 91.9\pmstd{1.8}
& 100.5\pmstd{1.5}
& ---
& 123.8\pmstd{2.5} \\
\midrule

\multirow{3}{*}{LOSO (no Norway)}
& TCN
& 88.6\pmstd{1.0}
& 92.2\pmstd{0.7}
& 163.9\pmstd{0.8}
& --- \\

& LSTM + SDF
& 86.6\pmstd{1.0}
& 95.6\pmstd{0.2}
& 142.1\pmstd{0.8}
& --- \\

& LSTM + G-Nbr Attn + SDF
& 90.4\pmstd{0.8}
& 97.6\pmstd{0.9}
& 149.4\pmstd{2.4}
& --- \\
\bottomrule
\end{tabular}%
}
\end{table}

\section{Conclusion}
\label{sec:conclusion}

We presented EnvShip, a unified framework for context-aware
and cross-region vessel trajectory forecasting. EnvShip processes
four public AIS sources through a shared pipeline and provides two
multi-horizon tracks with vessel-disjoint splits, sample-aligned
environmental, interaction, and meteorological context, and
transparent quality-control metadata.

Evaluation across regions, horizons, scene types, and training-source
compositions shows that forecasting performance depends strongly on
the operating regime. Physics-based and trajectory-only models remain
competitive in regular and open-water settings, whereas environmental
context is most useful when geographic constraints are informative,
its preferred representation nevertheless varies across scenes.
Neighboring-vessel context provides clearer benefits in
interaction-rich subsets than in aggregate evaluation. Multi-source
training can improve geographically distinct targets, but mismatched
sources may introduce negative transfer. These findings motivate
scene-, interaction-, and source-stratified reporting rather than
relying on a single aggregate ranking. Remaining limitations include
regional and seasonal imbalance, sparse interaction-rich cases, and
the inherent incompleteness of public AIS data. The benchmark
pipeline, evaluation tools, and processed data are available through
\repolink\ and \datalink.



\balance
\bibliographystyle{ACM-Reference-Format}
\bibliography{ship_env,ship_env_cleaned}

\clearpage
\appendix

\section{Dataset Construction and Curation Details}
\label{app:dataset_curation}

This appendix specifies how the quality-screened regional cores are
converted into the released standard tracks. Source adapters,
record-level filtering, trajectory segmentation, interpolation, and
resampling follow the shared pipeline described in the main paper
and are not repeated here.

\subsection{Regional Cores and Track Eligibility}
\label{app:regional_cores}

A regional core is the set of quality-screened candidate windows
obtained before standard-track curation. Each candidate retains its
MMSI, continuous-segment identifier, anchor time, vessel class,
train/validation/test assignment, trajectory-difficulty tier, and
sample-level quality metadata. The regional cores support alternative
task construction, whereas the standard tracks provide fixed subsets
for controlled model comparison.

A Track~A candidate must contain 30 observed and 30 future positions,
corresponding to 10 minutes of observation and 10 minutes of
prediction. A Track~B candidate must contain 90 observed and
180 future positions, corresponding to 30 minutes of observation
and 60 minutes of prediction. Candidates with an incomplete
observation--prediction interval or a failed trajectory-level quality
check are excluded before quota allocation.

Table~\ref{tab:app_curation_parameters} summarizes the parameters
specific to standard-track curation and the recommended
OpenStreetMap audit.

\begin{table}[t]
\centering
\caption{Parameters used for standard-track curation and auditing.}
\label{tab:app_curation_parameters}
\scriptsize
\setlength{\tabcolsep}{3.5pt}
\renewcommand{\arraystretch}{1.05}
\begin{tabular}{p{2.65cm}p{4.05cm}}
\toprule
\textbf{Component} & \textbf{Setting} \\
\midrule
Split assignment
& Deterministic MMSI hash; 80/10/10
train/validation/test split \\
Difficulty allocation
& 35/40/25 easy/medium/hard \\
Training caps
& At most 55 windows per vessel and
12 per continuous segment \\
Validation/test caps
& At most 70 windows per vessel and
20 per continuous segment \\
Within-segment spacing
& Anchor-time separation of at least
$\Delta_{\min}$, defined in the released configuration \\
Recommended OSM criterion
& Maximum inland depth no greater than
30\,m and fewer than three consecutive inland points \\
\bottomrule
\end{tabular}
\end{table}

\subsection{Core-to-Track Curation}
\label{app:core_to_track}

Curation is performed independently within each source and split.
The vessel-disjoint assignment established before curation is
preserved throughout the procedure; samples are never reassigned
across train, validation, and test sets.
Table~\ref{tab:app_curation_flow} summarizes the selection process.

\begin{table*}[t]
\centering
\caption{Procedure for converting regional-core candidates into the
released standard tracks.}
\label{tab:app_curation_flow}
\scriptsize
\setlength{\tabcolsep}{4.0pt}
\renewcommand{\arraystretch}{1.08}
\begin{tabular}{p{0.65cm}p{2.75cm}p{8.7cm}p{3.2cm}}
\toprule
\textbf{Step} & \textbf{Stage} & \textbf{Rule} & \textbf{Output} \\
\midrule
1
& Protocol eligibility
& Retain candidates containing the complete Track~A or Track~B
observation--prediction interval and satisfying all trajectory-level
quality checks.
& Track-specific candidate pools \\

2
& Split preservation
& Preserve the deterministic MMSI-based split assignment so that
no vessel occurs in more than one split.
& Vessel-disjoint source--split pools \\

3
& Stratum construction
& Group candidates by source, split, vessel class, and difficulty
tier, and determine quotas subject to candidate availability.
& Per-stratum candidate pools and quotas \\

4
& Quality ordering
& Order candidates within each stratum using interpolation ratio,
displacement, path efficiency, speed variation, and temporal
continuity.
& Ranked candidates \\

5
& Redundancy control
& Enforce per-vessel and per-segment caps and require
$|a_i-a_j|\geq\Delta_{\min}$ for selected windows from the same
continuous segment.
& Redundancy-controlled candidates \\

6
& Quota completion
& Select ranked candidates until the corresponding source--split
quota is filled or the eligible pool is exhausted.
& Standard-track samples \\

7
& Context and audit
& Link context records through the stable sample identifier and
compute the OpenStreetMap temporal-consistency diagnostics.
& Complete set and recommended subset \\
\bottomrule
\end{tabular}
\end{table*}

Vessel-class allocation uses the shared taxonomy defined in the main
paper. Class proportions are controlled at the benchmark level rather
than imposed independently on every region. This preserves regional
differences in fleet composition and avoids exhausting vessel classes
that are uncommon in a particular source.

Difficulty tiers are defined by the turning-based score in
Eq.~\ref{eq:difficulty}. Candidates are classified as easy
($D_i<5$), medium ($5\leq D_i<12$), or hard ($D_i\geq12$), with a
target allocation of 35/40/25. Because the score depends on the
complete observed-and-future trajectory, it is used only for dataset
curation and stratified evaluation and is never provided to a
forecasting model.

The vessel and segment caps reduce repeated sampling of similar
motion from the same vessel or continuous segment. For any two
selected windows $i$ and $j$ from the same segment,

\[
|a_i-a_j|\geq\Delta_{\min},
\]

where $a_i$ and $a_j$ are their anchor times. The value of
$\Delta_{\min}$ for each curation configuration is stored in the
version-controlled release files.

Track~A reaches the fixed source allocation of 150{,}000 samples
from DMA and 60{,}000 samples from each of NOAA, Piraeus, and
Norway. Track~B contains 106{,}857 samples. Its regional imbalance
reflects the availability of continuous 270-position windows rather
than post-hoc filtering based on model performance. Exact
source--split counts are reported in Table~\ref{tab:source_counts}.

\subsection{OpenStreetMap Temporal-Consistency Audit}
\label{app:temporal_consistency}

OpenStreetMap geometry may postdate the corresponding AIS
observations. Coastal infrastructure added after the observation
period, such as reclaimed land, piers, and breakwaters, can therefore
cause historically valid vessel trajectories to appear inland relative
to a later map snapshot.

Let $d_{i,t}^{\mathrm{in}}\geq0$ denote the inland depth of trajectory
point $t$ in sample $i$, as derived from the shoreline signed-distance
field, with zero assigned to points that are not inland. The audit
records

\[
d_i^{\max}
=
\max_t d_{i,t}^{\mathrm{in}},
\qquad
n_i^{\mathrm{in}}
=
\sum_t
\mathbf{1}\!\left[d_{i,t}^{\mathrm{in}}>0\right],
\qquad
\ell_i^{\mathrm{in}}
=
\operatorname{LongestRun}_t
\mathbf{1}\!\left[d_{i,t}^{\mathrm{in}}>0\right].
\]

Here, $d_i^{\max}$ is the maximum inland penetration,
$n_i^{\mathrm{in}}$ is the total number of inland points, and
$\ell_i^{\mathrm{in}}$ is the longest consecutive run of inland
points. A sample belongs to the recommended temporally consistent
subset when

\[
d_i^{\max}\leq30\,\mathrm{m}
\qquad\text{and}\qquad
\ell_i^{\mathrm{in}}<3.
\]

The audit does not remove samples from the release. The complete
set, the pass/fail flag, and all diagnostic values are provided so
that users can apply alternative criteria. The recommended criterion
retains 326{,}316 of the 330{,}000 Track~A samples and 99{,}191 of
the 106{,}857 Track~B samples.

For Piraeus, whose AIS observations were collected in 2019,
environmental context and the corresponding audit use a pinned
OpenStreetMap snapshot dated 2020-01-01.

The target trajectory and split assignment are fixed before context
records are attached. Environmental context, neighboring-vessel
features, and meteorological attributes use only information
available at or before the anchor time. The temporal-consistency
audit examines the complete trajectory only to generate quality
diagnostics and the recommended-subset flag; these values are not
used as forecasting inputs. Consequently, neither context
construction nor auditing changes split membership or introduces
prediction-interval information into the model inputs.

\section{Reference Models and Context Packages}
\label{app:baselines_context}

This appendix provides implementation details for the reference
models and defines the released environmental, neighboring-vessel,
and meteorological context records. The common forecasting protocol,
training procedure, and evaluation metrics are described in the main
paper and are not repeated here.

\subsection{Kinematic and Classical Baselines}
\label{app:baseline_definitions}

Let $\mathbf{y}_{T_h}$ denote the final observed position,
$\Delta t=20$\,s the sampling interval, and
$t=1,\ldots,T_f$ a future step.

\paragraph{Constant Velocity.}
The velocity is estimated from the final $M=3$ observed
displacements:

\[
\bar{\mathbf{v}}
=
\frac{1}{M}
\sum_{k=T_h-M}^{T_h-1}
\frac{\mathbf{y}_{k+1}-\mathbf{y}_{k}}{\Delta t},
\qquad
\hat{\mathbf{y}}_{T_h+t}
=
\mathbf{y}_{T_h}
+t\Delta t\,\bar{\mathbf{v}}.
\]

\paragraph{Dead Reckoning.}
The final reported speed over ground is converted to meters per
second and combined with the final course over ground expressed in
the target-centered local frame:

\[
\mathbf{v}^{\mathrm{DR}}
=
s_{T_h}\mathbf{u}(\psi_{T_h}),
\qquad
\hat{\mathbf{y}}_{T_h+t}
=
\mathbf{y}_{T_h}
+t\Delta t\,\mathbf{v}^{\mathrm{DR}},
\]

where $s_{T_h}$ and $\psi_{T_h}$ are the final observed SOG and COG,
and $\mathbf{u}(\psi)$ is the corresponding unit direction vector.

\paragraph{Additional released baselines.}
The released baseline grid also includes Constant Acceleration,
Random Forest, and XGBoost. Constant Acceleration fits independent
quadratic functions to the final five observed $x$- and
$y$-coordinates by least squares:

\[
\mathbf{y}(\tau)
=
\mathbf{a}
+\mathbf{b}\tau
+\frac{1}{2}\mathbf{c}\tau^2,
\]

and evaluates the fitted functions at future timestamps. Random
Forest and XGBoost receive flattened trajectory and motion features.
A separate two-dimensional regressor predicts each future step, so
their outputs are non-autoregressive.

\subsection{Neural and Context-Aware Implementations}
\label{app:model_implementations}

\paragraph{Trajectory-only models.}
The LSTM, GRU, and bidirectional LSTM encoders use two recurrent
layers with 256 hidden units. Their autoregressive decoders predict
future displacements, which are accumulated from the final observed
position. The TCN contains eight dilated causal-convolution layers
with 128 channels and kernel size 3, followed by a full-horizon
prediction head. Transformer-NAR encodes the complete observation
sequence and predicts the full future sequence in one forward pass.

\paragraph{Neighbor-aware models.}
A shared multilayer perceptron embeds each valid neighboring vessel.
Neighbor Pool aggregates the valid embeddings into a single
permutation-invariant context vector. Gated Neighbor Attention
instead applies validity-masked cross-attention at each decoder step
using the neighbor embeddings together with CPA, TCPA, and
relative-speed attributes. A neighbor-count gate suppresses this
pathway when no valid neighbor is available.

\paragraph{Environmental models.}
The binary-raster and SDF models use the same three-stage CNN with
feature widths 16, 32, and 64, producing a
$64\times16\times16$ feature map. Global-pooling variants compress
this map into one context vector. Spatial-attention variants retain
the $16\times16$ spatial grid as 256 tokens and query these tokens
at each decoder step.

The descriptor model applies feature-wise linear modulation to the
decoder state:

\[
\operatorname{FiLM}
(\mathbf{h}_t,\mathbf{d})
=
\boldsymbol{\gamma}(\mathbf{d})
\odot
\mathbf{h}_t
+
\boldsymbol{\beta}(\mathbf{d}),
\]

where $\mathbf{d}$ is the environmental descriptor and
$\mathbf{h}_t$ is the decoder state.

\paragraph{Joint-context models.}
The descriptor-based joint model combines neighboring-vessel
attention with descriptor FiLM. The SDF-based joint model combines
gated neighboring-vessel attention with spatial attention over the
SDF feature map. Matched variants use the same trajectory backbone,
prediction target, normalization procedure, and training protocol,
so their differences are limited to the supplied context pathway.

\subsection{Environmental Context Package}
\label{app:environment_schema}

Environmental context is constructed once at the anchor for both
tracks. Each sample uses an axis-aligned
$10\,\mathrm{km}\times10\,\mathrm{km}$ crop in the target-centered
local frame, represented on a $128\times128$ grid. The spatial
resolution is therefore 78.125\,m per pixel.

\begin{table}[t]
\centering
\caption{Environmental context representations}
\label{tab:app_environment_schema}
\scriptsize
\setlength{\tabcolsep}{3.0pt}
\renewcommand{\arraystretch}{1.05}
\begin{tabular}{p{2.35cm}p{1.45cm}p{3.35cm}}
\toprule
\textbf{Representation} & \textbf{Shape} & \textbf{Definition} \\
\midrule
Land mask
& $128\times128$
& Pixels classified as land \\

Water mask
& $128\times128$
& Pixels classified as water \\

Navigable-water mask
& $128\times128$
& Water pixels associated with mapped navigable geometry \\

Natural-boundary mask
& $128\times128$
& Coastlines and other natural boundaries \\

Man-made-boundary mask
& $128\times128$
& Quays, piers, waterfronts, and related structures \\

Barrier mask
& $128\times128$
& Breakwaters and other mapped barriers \\

Shoreline SDF
& $128\times128$
& Signed distance to the nearest shoreline \\

Navigability SDF
& $128\times128$
& Signed distance to the nearest navigable-region boundary \\

Scene descriptor
& $14$
& Scene encoding and scalar geometric and quality attributes \\

Vector geometry
& Variable
& Clipped shoreline and maritime-feature polylines \\
\bottomrule
\end{tabular}
\end{table}

For a boundary set $\partial\Omega$, the signed-distance value at
location $\mathbf{p}$ is

\[
d(\mathbf{p})
=
s(\mathbf{p})
\min_{\mathbf{q}\in\partial\Omega}
\left\|
\mathbf{p}-\mathbf{q}
\right\|_2,
\]

where $s(\mathbf{p})\in\{-1,+1\}$ denotes the side of the boundary
under the released sign convention. The shoreline and navigability
SDFs use shoreline and navigable-region boundaries, respectively.

The 14-dimensional environmental descriptor follows the fixed
ordering in Table~\ref{tab:app_descriptor}.

\begin{table}[t]
\centering
\caption{Semantic organization of the environmental descriptor}
\label{tab:app_descriptor}
\scriptsize
\setlength{\tabcolsep}{3.4pt}
\renewcommand{\arraystretch}{1.05}
\begin{tabular}{p{2.25cm}p{0.75cm}p{4.15cm}}
\toprule
\textbf{Group} & \textbf{Dim.} & \textbf{Content} \\
\midrule
Scene encoding
& 5
& Indicators for open-water, nearshore, harbor, complex-environment,
and constrained scenes \\

Area ratios
& 2
& Water and navigable-water proportions within the crop \\

Nearest distances
& 3
& Distances to shoreline, navigable-region boundary, and barrier
geometry \\

Boundary densities
& 3
& Natural-, man-made-, and barrier-boundary densities \\

Context quality
& 1
& Completeness or reliability of the constructed context \\
\midrule
Total
& 14
& Fixed ordering defined in the released schema \\
\bottomrule
\end{tabular}
\end{table}

Binary scene indicators are used directly. Continuous descriptor
channels are standardized using statistics estimated only from the
corresponding training pool:

\[
\widetilde{z}
=
\frac{
z-\mu_{\mathrm{train}}
}{
\sigma_{\mathrm{train}}+\varepsilon
}.
\]

The same statistics are applied unchanged to validation and test
samples.

\subsection{Neighboring-Vessel Context}
\label{app:neighbor_schema}

At the anchor time, valid surrounding vessels are ordered by
Euclidean distance to the target. The nearest
$K_{\max}=10$ vessels within 3\,km are retained. Samples with fewer
neighbors are padded to the fixed size, with padded entries identified
by a binary validity mask.

\begin{table}[t]
\centering
\caption{Neighboring-vessel representations with
$K_{\max}=10$}
\label{tab:app_neighbor_schema}
\scriptsize
\setlength{\tabcolsep}{3.0pt}
\renewcommand{\arraystretch}{1.05}
\begin{tabular}{p{2.25cm}p{1.65cm}p{3.2cm}}
\toprule
\textbf{Input} & \textbf{Shape} & \textbf{Content} \\
\midrule
Relative history
& $K_{\max}\times T_h\times2$
& Neighbor positions relative to the target history \\

Relative velocity
& $K_{\max}\times2$
& $\Delta v_x$ and $\Delta v_y$ at the anchor \\

Basic features
& $K_{\max}\times5$
& $\Delta x$, $\Delta y$, $\Delta v_x$, $\Delta v_y$, and distance
$d$ \\

Extended features
& $K_{\max}\times8$
& Basic features, CPA, TCPA, and relative-speed magnitude
$\|\Delta\mathbf{v}\|_2$ \\

Validity mask
& $K_{\max}$
& One for a valid neighbor and zero for padding \\
\bottomrule
\end{tabular}
\end{table}

Let $\mathbf{p}_0$ and $\mathbf{v}_0$ denote the target position and
velocity at the anchor, and let $\mathbf{p}_j$ and $\mathbf{v}_j$
denote the corresponding state of neighbor $j$. The relative position
and velocity are

\[
\mathbf{r}_j
=
\mathbf{p}_j-\mathbf{p}_0,
\qquad
\mathbf{u}_j
=
\mathbf{v}_j-\mathbf{v}_0.
\]

Under a constant-relative-velocity assumption, Time to Closest Point
of Approach and Closest Point of Approach are computed as

\[
\mathrm{TCPA}_j
=
-
\frac{
\mathbf{r}_j^\top\mathbf{u}_j
}{
\|\mathbf{u}_j\|_2^2+\varepsilon
},
\qquad
\mathrm{CPA}_j
=
\left\|
\mathbf{r}_j
+
\mathrm{TCPA}_j\mathbf{u}_j
\right\|_2.
\]

Both quantities use only states available at the anchor. The released
implementation applies the configured clipping and normalization
before these attributes are supplied to a model. No state from the
prediction interval is used.

The validity mask is defined as

\[
m_{i,j}
=
\begin{cases}
1, & \text{if neighbor }j\text{ is present and valid},\\
0, & \text{if the entry is padded or unavailable}.
\end{cases}
\]

Pooling and attention operations exclude entries for which
$m_{i,j}=0$.

\subsection{Weather, Sea-State, and Traffic Attributes}
\label{app:weather_availability}

Each sample is matched at its anchor location and time to six ERA5
weather attributes and four ECMWF wave and sea-state attributes.
The exact field names, units, and ordering are defined in the released
schema. Nearest-port information and three indicators describing
nearby fairways or traffic-separation schemes are derived separately
from maritime and OpenStreetMap geometry.

All four AIS sources use the same matching procedure. Availability is
determined at the sample level: an attribute is marked as unavailable
when the requested time--location pair lies outside usable
external-product coverage or fails the corresponding matching-quality
check. Missing values are represented by availability masks and are
not numerically imputed.

Per-source and per-attribute availability statistics are included in
the released metadata. This allows model implementations to
distinguish an unavailable attribute from a physically observed zero.

\subsection{Environmental Spatial Support}
\label{app:coverage_definition}

The environmental crop is fixed at the anchor and remains unchanged
throughout prediction. Because the crop is a square, a future
trajectory is fully covered only when all of its positions remain
within the $\pm5$\,km bounds on both local axes. Define

\[
C_i
=
\mathbf{1}
\left[
\max_{1\leq t\leq T_f}|x_{i,t}|
\leq5\,\mathrm{km}
\ \land\
\max_{1\leq t\leq T_f}|y_{i,t}|
\leq5\,\mathrm{km}
\right].
\]

A sample is fully covered when $C_i=1$. The crop has a 5\,km
half-width and an anchor-to-corner distance of

\[
5\sqrt{2}
\approx
7.07\,\mathrm{km}.
\]

Endpoint displacement alone is therefore not an exact coverage test:
a point between 5 and 7.07\,km from the anchor may lie either inside
or outside the square depending on its direction, and a trajectory
may leave the crop before its endpoint. The indicator $C_i$ should
therefore be used for sample-level coverage analysis.

The 3\,km neighboring-vessel radius serves a different purpose. It
defines neighbor selection at the anchor and is not a spatial
coverage boundary for the target's future trajectory.

\section{Additional Analyses and Release Details}
\label{app:additional_results}

This appendix provides additional analyses of environmental context
at different prediction horizons, qualitative context examples, and
a summary of the released experimental records.

\subsection{Horizon-Dependent Utility of Environmental Context}
\label{app:context_utility}

Long-horizon trajectories have substantially greater spatial extent
than short-horizon trajectories. Table~\ref{tab:app_displacement_extent}
reports the displacement of the future endpoint from the anchor.
For DMA, the median displacement increases from 3.33\,km on
Track~A to 5.75\,km on Track~B. The corresponding Track~B median
for NOAA is 8.09\,km.

\begin{table}[t]
\centering
\caption{Future endpoint displacement from the anchor}
\label{tab:app_displacement_extent}
\scriptsize
\setlength{\tabcolsep}{5.0pt}
\renewcommand{\arraystretch}{1.05}
\begin{tabular}{lrrr}
\toprule
\textbf{Track / region}
& \textbf{Horizon}
& \textbf{Median}
& \textbf{P90} \\
& \textbf{(min)}
& \textbf{(km)}
& \textbf{(km)} \\
\midrule
Track~A / DMA
& 10 & 3.33 & 4.68 \\
Track~B / DMA
& 60 & 5.75 & 10.93 \\
Track~B / NOAA
& 60 & 8.09 & 11.47 \\
\bottomrule
\end{tabular}
\end{table}

\begin{figure}[t]
    \centering
    \includegraphics[width=0.80\columnwidth]{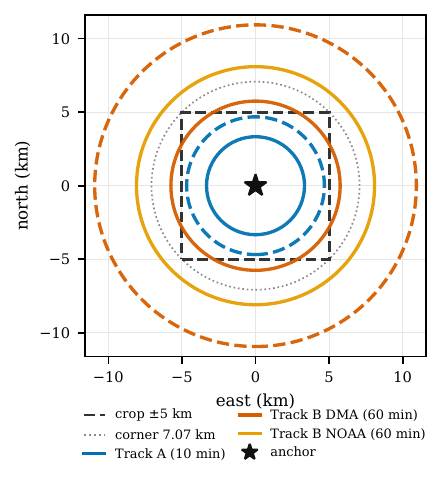}
    \caption{Fixed anchor-centered environmental crop and radial
endpoint-displacement quantiles. The dashed square denotes the
$\pm5$\,km crop, and the circles denote median and 90th-percentile
endpoint distances.
These radial summaries illustrate the greater
spatial extent of Track~B 
}
    \label{fig:app_spatial_coverage}
\end{figure}

The DMA Track~A 90th-percentile endpoint displacement is below the
5\,km crop half-width. In contrast, the Track~B 90th percentiles for
DMA and NOAA exceed the 7.07\,km anchor-to-corner distance, and the
NOAA Track~B median also exceeds this distance
(Fig.~\ref{fig:app_spatial_coverage}). These endpoint statistics
illustrate the greater spatial extent of Track~B but do not constitute
an exact crop-coverage test. Exact sample-level coverage depends on
all future positions and is defined in
Appendix~\ref{app:coverage_definition}.

To examine how environmental utility changes with the prediction
horizon, we compare the SDF model with its trajectory-only LSTM
backbone. Table~\ref{tab:app_context_horizon} reports prefix ADE on
DMA Track~B over trajectories ending at 3, 6, 10, and 60 minutes,
using the same five-seed protocol as the full-horizon evaluation.

The relatively large early-horizon errors of both recurrent models
reflect their optimization for the complete 180-step autoregressive
forecast rather than a data or metric inconsistency. Because
long-range residuals are substantially larger, the full-horizon
objective can trade near-term accuracy for performance over the
complete rollout. Within this matched comparison, SDF conditioning
reduces ADE by 9.3--12.7\% over the first 10 minutes, compared with
2.2\% over the full 60-minute horizon. The absolute improvement
remains at 60 minutes, but its relative contribution decreases as
forecasting errors accumulate. Together with the endpoint-displacement
statistics, this result suggests that fixed anchor-centered context is
most directly useful at earlier horizons, while its effect is diluted
over long autoregressive rollouts.

\begin{table}[t]
\centering
\caption{Horizon-wise prefix ADE (m) on the DMA Track~B test set
($N=6{,}000$) for models trained on the full 60-minute forecasting
task, averaged over five seeds.
$\Delta\mathrm{ADE}
=\mathrm{ADE}_{\mathrm{SDF}}-\mathrm{ADE}_{\mathrm{LSTM}}$,
and gain denotes the relative ADE reduction over the LSTM baseline}
\label{tab:app_context_horizon}
\scriptsize
\setlength{\tabcolsep}{3.8pt}
\renewcommand{\arraystretch}{1.05}
\begin{tabular}{lrrrr}
\toprule
\textbf{Horizon}
& \textbf{LSTM}
& \textbf{LSTM + SDF}
& \textbf{$\Delta$ADE}
& \textbf{Gain} \\
\midrule
3 min
& 86.3 & 77.2 & $-9.1$ & 10.6\% \\
6 min
& 88.1 & 76.9 & $-11.2$ & 12.7\% \\
10 min
& 103.1 & 93.5 & $-9.6$ & 9.3\% \\
60 min
& 658.5 & 643.7 & $-14.8$ & 2.2\% \\
\bottomrule
\end{tabular}
\end{table}

This analysis does not identify limited spatial support as the sole
cause of the reduced relative gain. Regional scene composition,
trajectory complexity, error accumulation, and optimization
variability may also contribute. The fixed crop represents a
trade-off among spatial extent, resolution, storage, and computation.
Multi-scale environmental representations or dynamically updated
context provide possible directions for long-horizon forecasting.

\subsection{Qualitative Context Examples}
\label{app:qualitative_context}

\begin{figure*}[t]
    \centering
    \includegraphics[width=0.85\textwidth]
    {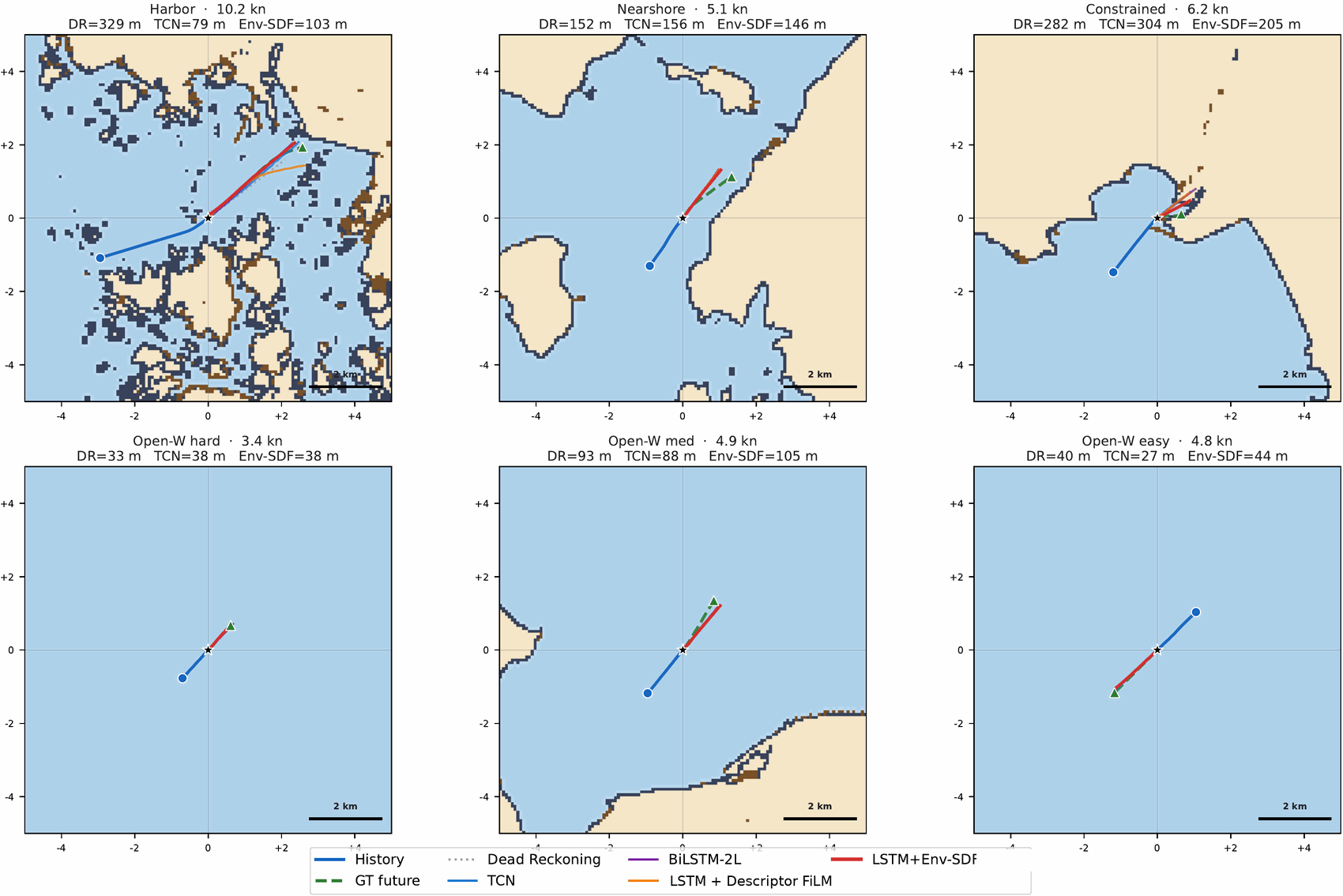}
    \caption{Representative deterministic Track~A predictions across
    maritime scenes. Observed and ground-truth trajectories are shown
    together with kinematic, trajectory-only, descriptor-based, and
    SDF-based baselines}
    \label{fig:app_prediction_examples}
\end{figure*}

Figure~\ref{fig:app_prediction_examples} shows representative
deterministic Track~A predictions in different maritime scenes.
The examples illustrate qualitative differences among kinematic,
trajectory-only, descriptor-based, and SDF-based models and are not
used for quantitative model comparison.

\begin{figure*}[t]
    \centering
    \includegraphics[width=0.62\textwidth]
    {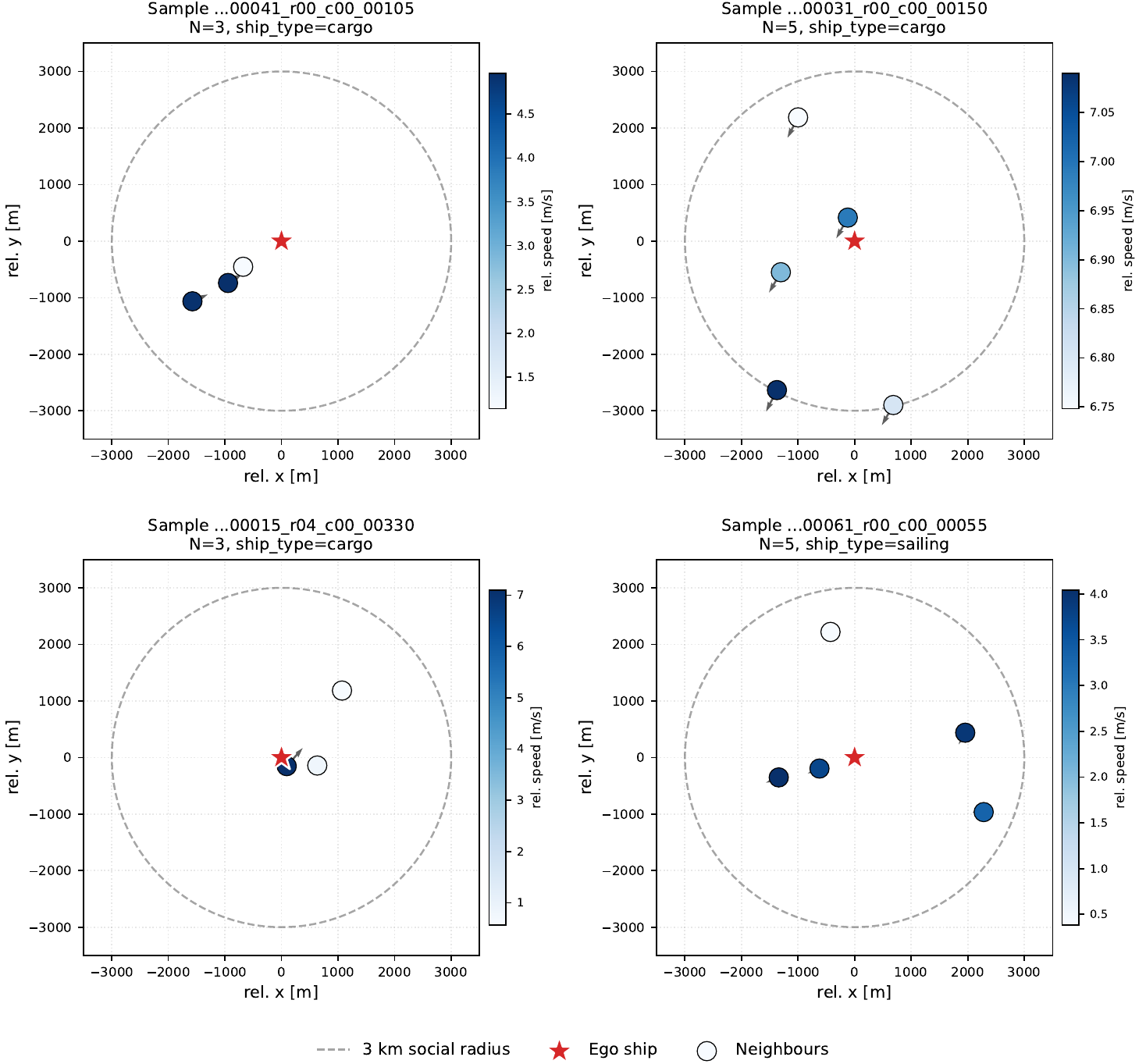}
    \caption{Neighboring-vessel context for four DMA test samples.
    The panels show the 3\,km selection radius, target and neighboring
    vessels, and anchor-time relative-speed attributes}
    \label{fig:app_social_encoding}
\end{figure*}

Figure~\ref{fig:app_social_encoding} illustrates the anchor-time
representation used to construct neighboring-vessel context.
Positions are expressed in the target-centered local frame, and
marker intensity encodes relative-speed magnitude. CPA and TCPA are
computed from the same anchor-time relative positions and velocities,
as defined in Appendix~\ref{app:neighbor_schema}. The presence of a nearby vessel does not by itself indicate a
safety-critical encounter. The released interaction attributes are
intended as forecasting context rather than as certified
collision-risk estimates.

\subsection{Released Experimental Records}
\label{app:result_records}

The release includes the complete model--region result grid,
intermediate-horizon metrics for both tracks, difficulty-, scene-,
and neighbor-stratified evaluations, and the available per-seed
records. Each result record identifies the dataset version, track,
source region, training pool, evaluation split, temporal-consistency
subset, model configuration, random seed, and selected checkpoint.

These fields distinguish experiments conducted under different
source compositions, quality-control settings, and statistical
protocols. The associated configuration files define model and
training parameters required to reproduce each reported result.

\subsection{Release, Licensing, and Intended Use}
\label{app:access_licensing}

The public release contains the benchmark trajectories,
sample-aligned context, quality-control metadata, construction and
evaluation code, and experimental records. A stable sample identifier
links each trajectory to its context records, quality flags, and split
assignment without requiring additional data construction.

Because EnvShip combines multiple upstream resources, license and
attribution requirements are documented separately for each source.
The release includes a per-file license mapping and the corresponding
attribution notices for AIS records, OpenStreetMap-derived
environmental layers, and meteorological and sea-state products.
Users are responsible for following the applicable upstream terms
when redistributing the data or derived products.

EnvShip is intended for research on vessel trajectory forecasting,
context modeling, and cross-region evaluation. The release does not
add vessel ownership, operator, crew, or other identity information
to the source AIS records. It should not be used to re-identify or
monitor individual vessels or crews.
Dataset versions, schema changes, and known issues are documented
through the public release and its versioned changelog.

\end{document}